\documentclass[10pt,twocolumn,letterpaper]{article}
\usepackage{cvpr}     
\graphicspath{figures}
\usepackage{tabularx}
\usepackage{booktabs}
\usepackage{dcolumn}
\usepackage{amsfonts}
\usepackage{subcaption}
\usepackage{booktabs}
\usepackage{threeparttable}
\usepackage{multirow}
\usepackage{mathrsfs}

\usepackage{graphicx}
\usepackage{amsmath}
\usepackage{amssymb}
\usepackage{booktabs}
\usepackage{algorithm}
\usepackage{algorithmic}

\usepackage[pagebackref,breaklinks,colorlinks,allcolors=cvprblue]{hyperref}
\usepackage[capitalize]{cleveref}
\crefname{section}{Sec.}{Secs.}
\Crefname{section}{Section}{Sections}
\Crefname{table}{Table}{Tables}
\crefname{table}{Tab.}{Tabs.}

\usepackage{cvpr}      
\definecolor{cvprblue}{rgb}{0.21,0.49,0.74}

\def\paperID{9629} 
\def\confName{CVPR}
\def\confYear{2025}

\title{{Consistent Diffusion}: Denoising Diffusion Model with Data-Consistent Training for Image Restoration}

\author{
Xinlong Cheng, Tiantian Cao, Guoan Cheng, Bangxuan Huang, Xinghan Tian, Ye Wang, Xiaoyu He, \\
\and 
Weixin Li, Tianfan Xue, Xuan Dong* 
}

\begin{document}
\maketitle

\begin{abstract}

In this work, we address the limitations of denoising diffusion models (DDMs) in image restoration tasks, particularly the shape and color distortions that can compromise image quality. While DDMs have demonstrated a promising performance in many applications such as text-to-image synthesis, their effectiveness in image restoration is often hindered by shape and color distortions. We observe that these issues arise from inconsistencies between the training and testing data used by DDMs. Based on our observation, we propose a novel training method, named data-consistent training, which allows the DDMs to access images with accumulated errors during training, thereby ensuring the model to learn to correct these errors. Experimental results show that, across five image restoration tasks, our method has significant improvements over state-of-the-art methods while effectively minimizing distortions and preserving image fidelity.
\end{abstract}

\section{Introduction}
\label{sec:intro}

The denoising diffusion model (DDM) has demonstrated promising results in various tasks, e.g. text-to-image synthesis, image style transfer, etc. To expand the applications of DDMs, some studies have explored its use in general image restoration tasks, including ResShift \cite{yue2024resshift}, Diff-Plugin \cite{liu2024diff}, DiffUIR \cite{zheng2024selective}, etc.
However, the results of DDMs often exhibit shape and color distortions. While these distortions may be acceptable for entertainment tasks, they significantly degrade image quality in restoration tasks, such as denoising, super-resolution, and dehazing, where high fidelity to the input images is crucial for practical applications like in-camera ISP and image editing software.

\begin{figure}[t] 
    \centering
    \begin{minipage}[ht]{0.47\linewidth} 
        \centering
        \includegraphics[width=\linewidth, height=0.8\linewidth]{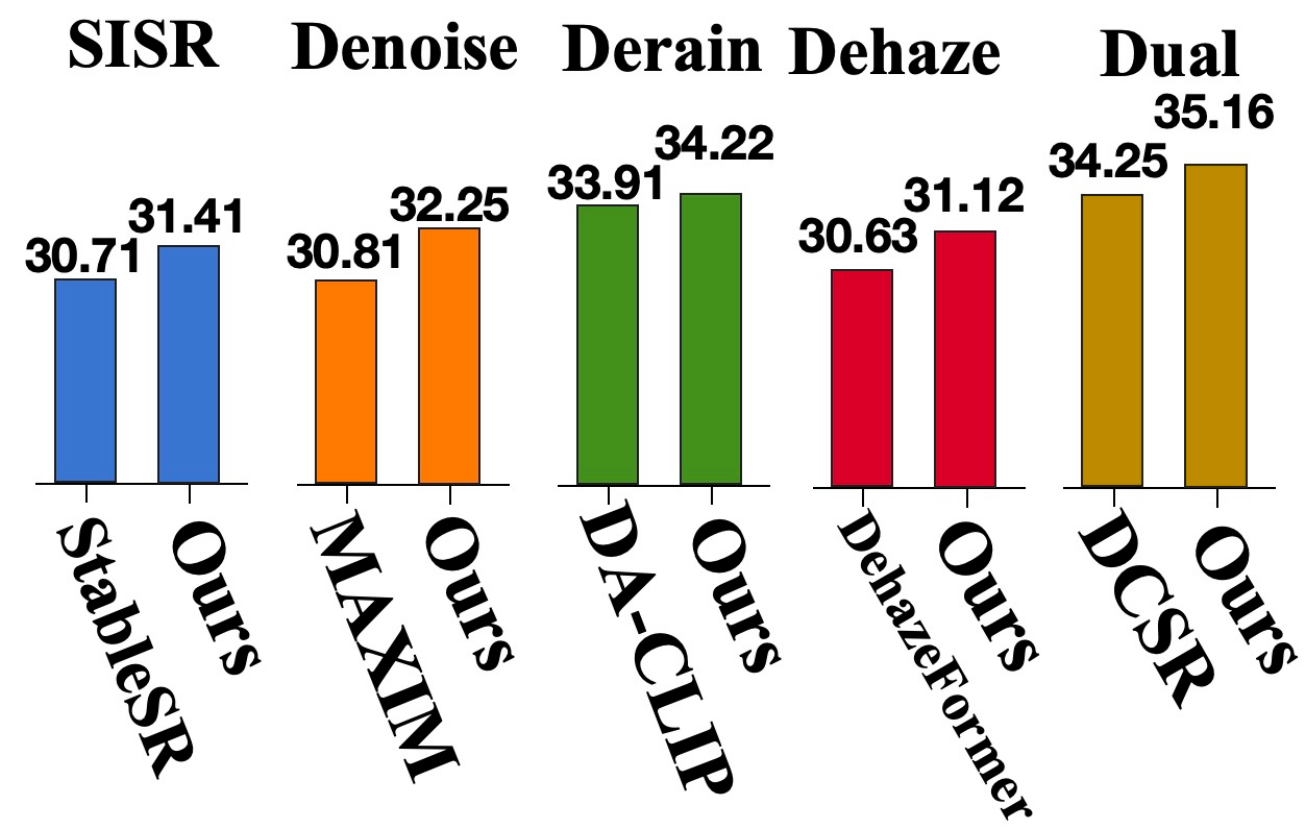} 
        \vspace{-5pt}  
        \subcaption{PSNR (dB) values on five tasks.}
        \vspace{-5pt} 
    \end{minipage}%
    \hspace{5pt} 
    \begin{minipage}[ht]{0.47\linewidth} 
        \centering
        \includegraphics[width=\linewidth, height=0.8\linewidth]{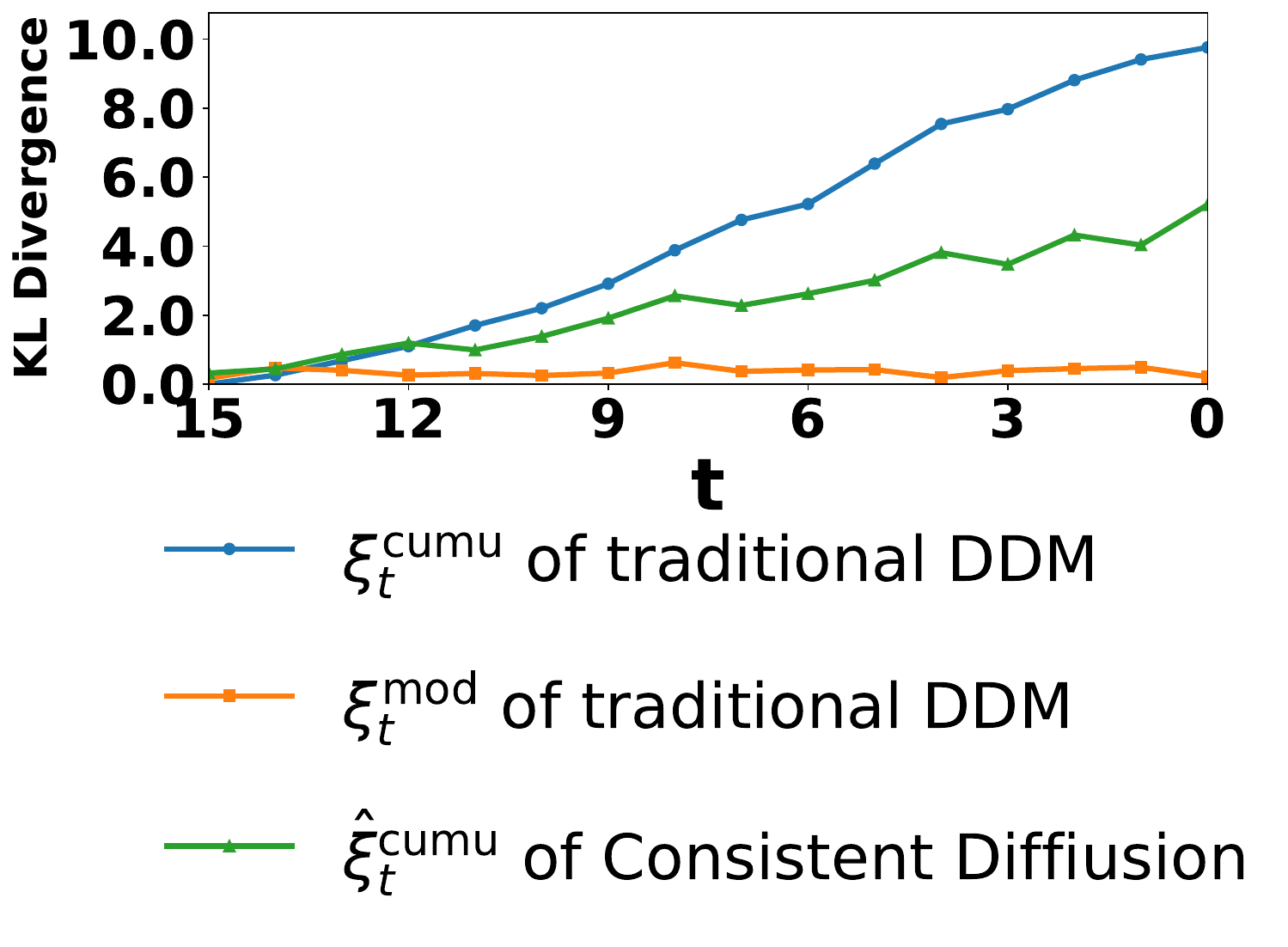} 
        \vspace{-5pt} 
        \subcaption{Modular and cumulative errors.}
        \vspace{-5pt} 
    \end{minipage}
    \vspace{-1pt}  
    \caption{(a) shows PSNR values of the best comparison method and our consistent diffusion on five image restoration tasks: single image super-resolution (SISR), denoise, derain, dehaze, and dual camera super-resolution (Dual). (b) shows that in traditional DDM, the modular error ${\xi}_{t}^{\rm{mod}}$ during training remains low across each iteration $t$, while the cumulative error ${\xi}_{t}^{\rm{cumu}}$ continuously grows with each testing iteration $t$. By comparison, in our consistent diffusion, the cumulative error ${\hat{\xi}}_{t}^{\rm{cumu}}$ during testing remains much lower than ${\xi}_{t}^{\rm{cumu}}$.}
    \label{fig:teasor}
\end{figure}

\begin{figure*}[ht]
\begin{minipage}[ht]{.99\linewidth}
\centering
\vspace{-5pt}  
\subfloat{\label{}\includegraphics[width=\textwidth]{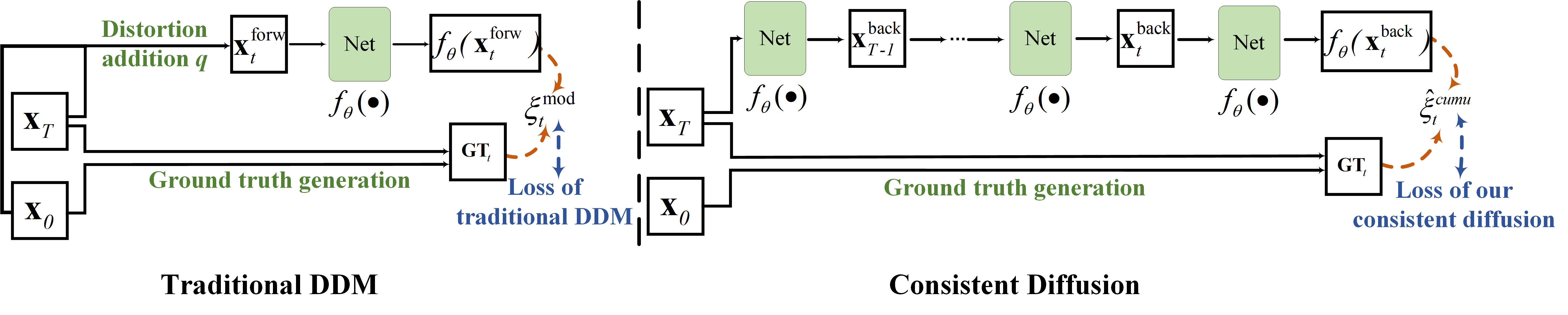}} 
\end{minipage}
\vspace{-5pt}
\captionsetup{belowskip=0pt} 
\caption{Pipelines of the training methods of traditional DDM vs. our consistent diffusion. At iteration $t$ of the training stage, traditional DDM employs a one-step forward process to obtain the input data ${\bf{x}}_t ^{\rm{train}}$, i.e. ${\bf{x}}_t ^{\rm{train}}={\bf{x}}_t^{{\rm{forw}}} \sim{q({{\bf{x}}_t}|{{\bf{x}}_0})}$, where $q$ is the distortion addition operation. The training loss, which measures the quality of $f_{\theta} ({\bf{x}}_t^{{\rm{forw}}})$, is optimizing the modular error ${\xi}_{t}^{\rm{mod}}$ of the core network while failing to take the input cumulative error into consideration. $f_{\theta}$ is the processing of the core network with the parameter $\theta$. In contrast, our consistent diffusion proposes the data-consistent training. It utilizes the multi-step backward process, which is consistent with that in the testing stage, to generate the input data ${\bf{x}}_t ^{\rm{train}}$ at iteration $t$, i.e. ${\bf{x}}_t ^{\rm{train}}={\bf{x}}_t^{{\rm{back}}} \sim{ {p_\theta }({{\bf{x}}_T})\prod\limits_{i= T}^{t+1} {{p_\theta }({{\bf{x}}_{i - 1}}|{{\bf{x}}_i})}}$. ${p_\theta }$ denotes the denoising process parameterized by $\theta$. The training loss, which measures the quality of $f_{\theta} ({\bf{x}}_t^{{\rm{back}}})$, is directly optimizing the cumulative error $\hat{{\xi}}_{t}^{\rm{cumu}}$ of the core network.}
\label{fig:pipeline}
\end{figure*}

To explore the sources of distortions of traditional DDMs, we analyzed their modular error during training and cumulative error during testing at different iterations. As explained in Fig. \ref{fig:pipeline} and \cite{li2023error}, the modular error reflects the error of restoring $x_t ^{\rm{train}}$ (the input data during training). $x_t ^{\rm{train}}$ is generated via the single-step forward processing at iteration $t$. The cumulative error reflects the error for sequentially running the core network from the first iteration $T$ to the intermediate iteration $t$. As shown in Fig. \ref{fig:teasor} (b), there exists a big gap between the modular error during training and the cumulative error during testing. This finding indicates that, while the training focuses on dealing with ${\bf{x}}_t ^{\rm{train}}$,
the discrepancies between ${\bf{x}}_t ^{\rm{train}}$ and ${\bf{x}}_t ^{\rm{test}}$ (the input data during testing) are overlooked. The discrepancies are accumulated to be significant through iterative processing during testing.

Based on our observations, we propose a data-consistent training method. As shown in Fig. \ref{fig:pipeline}, at iteration $t$ of the training stage, we align ${\bf{x}}_t ^{\rm{train}}$ with ${\bf{x}}_t ^{\rm{test}}$ by using backward processing to generate ${\bf{x}}_t ^{\rm{train}}$ at each iteration $t$. This approach minimizes the input differences at any iteration $t$ between training and testing. This ensures that the errors in the testing stage are fully considered during training, and allows the loss optimization in training to directly impact testing accuracy. In addition, to reduce the memory and computational cost of data-consistent training, we also provide an efficient version with a certain bias.

In our experiments, we select ResShift \cite{yue2024resshift} as the backbone DDM, one of the state-of-the-art (SOTA) models with minimal diffusion steps, and apply the proposed data-consistent training method to train the model. We evaluate the method on five popular image restoration tasks: single-image super-resolution, denoising, deraining, dehazing, and dual-camera super-resolution. Training is performed using a single NVIDIA A6000 GPU. The results in Fig. \ref{fig:teasor} and Sec. \ref{sec:experiments} show that our method achieves SOTA accuracy compared to other methods and produces high-fidelity outputs that effectively prevent color and shape distortions.

Our contributions can be summarized as follows. (1) We propose the data-consistent training method to mitigate error propagation during the iterative processing of DDM. (2)
The proposed method is applicable to any DDM backbone and can be used for various image restoration tasks. (3) Experiments across five image restoration tasks demonstrate that our method achieves SOTA accuracy while effectively preventing distortions.

\section{Related Work}
\label{sec:formatting}
\subsection{Restoration methods.}
Before the deep learning era, there exist various miles-stone non-deep solutions, e.g. sparse coding for image super resolution \cite{yang2010image}, non-local means for image denoising \cite{dabov2007image}, dark channel prior for image dehaze \cite{he2010single}, etc.

Among deep learning methods, CNN is widely used in different restoration tasks, including SISR \cite{lim2017enhanced,zhang2018image,lee2022local,zhang2024real}, denoising \cite{lefkimmiatis2018universal}, debluring \cite{nah2017deep,fang2023self,abuolaim2020defocus}, derain \cite{fu2017removing,Fu_2017,zhu2023learning}, dehaze \cite{cui2023irnext,ren2018gated,cui2024omni}, and dual-camera super resolution \cite{wang2021dual,zhang2022self,yue2024kedusr}.
With the increase of data and computation resources, Transformer becomes a more powerful backbone than CNN. Some works solve multiple restoration tasks, e.g. Pretrained-IPT \cite{chen2021pre}, Restormer \cite{zamir2022restormer}, U-Former \cite{wang2022uformer}, CSformer \cite{ye2023csformer}, SwinIR \cite{liang2021swinir}, PromptRestorer \cite{wang2023promptrestorer}, GRL \cite{li2023efficient}, and there also exist many specfically designed models for single tasks, including SISR \cite{liang2021swinir,chen2023activating,zhou2023srformer}, denoising \cite{xu2023cur}, debluring \cite{tsai2022stripformer,tsai2022banet,li2023efficient}, derain \cite{valanarasu2022transweather,xiao2022image,li2019heavy}, dehaze \cite{guo2022image,li2022physically,chen2024dea}, and reference-based super resolution \cite{yang2020learning}.

\subsection{Denoising Diffusion Models.}

Denoising Diffusion Models (DDMs) (e.g. DDPM \cite{ho2020denoising}, DDIM \cite{song2020denoising}, LDM \cite{rombach2022high}) provide a more complicated and powerful backbone in the computer vision field. In a wide range of tasks like text-to-image synthesis and image generation with multimodal inputs, DDMs have shown promising results, e.g. ControlNet \cite{zhang2023adding}, Uni-controlnet \cite{zhao2024uni}, smol-imagen \cite{saharia2022photorealistic}, Improved-NAT \cite{ni2024revisiting}, Splatter image \cite{szymanowicz2024splatter}, Diffusion-distiller \cite{salimans2022progressive}.

DDMs have also been studied to serve for image restoration. Some works try to design the model for various restoration tasks, e.g. GenerativeDiffusionPrior \cite{fei2023generative}, DiffIR \cite{xia2023diffir}, DiffPIR \cite{zhu2023denoising}, DiffUIR \cite{zheng2024selective}, Diff-Plugin \cite{liu2024diff}, FlowIE \cite{zhu2024flowie}, TextualDegRemoval \cite{lin2024improving}, and DTPM \cite{ye2024learning}. The other works focus on dealing with a single task, like single image super-resolution  \cite{yue2024resshift,wang2024sinsr,wang2024exploiting,wu2024seesr,yu2024scaling} and image debluring, e.g. DSR \cite{whang2022deblurring} and ID-Blau \cite{wu2024id}.

The existing DDM methods provide a variety of ways to revise the network structure and add pre-/post-processing subnets to improve the accuracy of DDMs for image restoration. We notice that the training and testing strategy of most of these DDMs follows the traditional DDPM and DDIM. However, the cumulative error through the iterative processing of DDMs during testing is not considered in the training stage, leading to uncontrolled errors which could decrease the fidelity and accuracy much a lot.

\begin{figure*}[ht] 
    \centering
    \begin{minipage}[t]{0.48\linewidth} 
        \centering
        \vspace*{-\topskip} 
        \includegraphics[width=\linewidth]{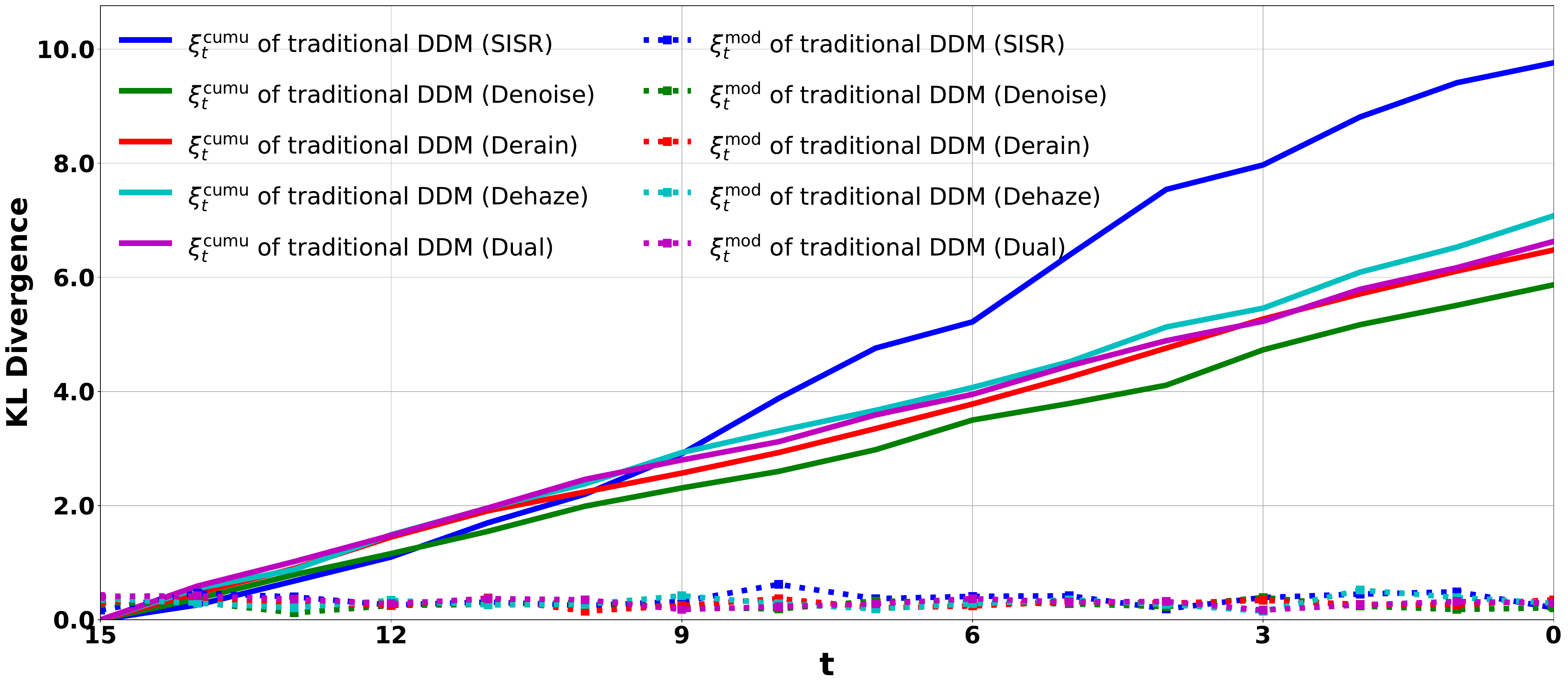} 
        \vspace{-5pt}  
        \subcaption{$\xi_{t}^{mod}$ and $\xi_{t}^{cumu}$ of traditional DDM on five restoration tasks.}
        \vspace{-5pt}  
    \end{minipage}%
    \hspace{5pt} 
    \begin{minipage}[t]{0.50\linewidth} 
        \centering
        \vspace*{-\topskip} 
        \includegraphics[width=\linewidth]{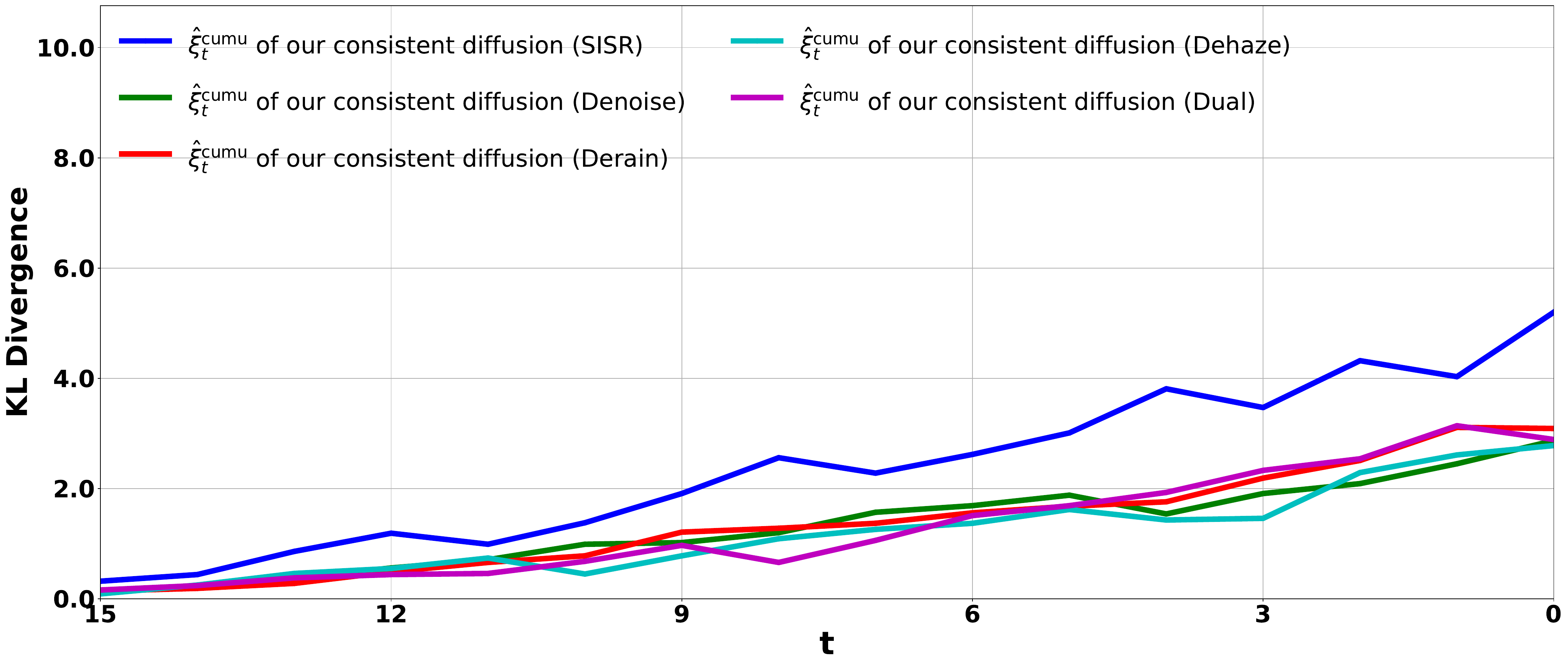} 
        \vspace{-5pt} 
        \subcaption{$\hat{\xi}_{t}^{cumu}$ of our consistent diffusion on five restoration tasks.}
        \vspace{-5pt}  
    \end{minipage}
    \vspace{-1pt}  
    \caption{(a) shows the statistics of traditional DDM across five tasks, revealing a persistent gap between the modular error $\xi_{t}^{mod}$, which is minimized by the loss function during training, and the cumulative error $\xi_{t}^{cumu}$, which represents the true error during testing. We use ResShift \cite{yue2024resshift} as the traditional DDM here. In contrast, (b) shows that the cumulative errors $\hat{\xi}_{t}^{cumu}$ of our consistent diffusion are significantly lower than cumulative errors $\xi_{t}^{cumu}$ of traditional DDM in (a), benefiting from our data-consistent training which directly optimizes for cumulative error reduction. Our consistent diffusion uses the backbone of ResShift with the proposed data-consistent training.}
    \label{fig:statistics}
\end{figure*}

\section{Preliminaries}

\subsection{Error analysis}
DDM is a sequence model which iteratively process the input data from low quality to high quality step-by-step to obtain the final output. As mentioned in \cite{li2023error} and explained in Fig. \ref{fig:statistics}, at iteration $t$, there exists the \textbf{modular error} and \textbf{cumulative error}.

\textbf{Modular error} $\xi_t^{{\rm{mod}}}$ measures the accuracy of the output of the core network given an input ${\bf{x}}_t$ at iteration $t$, i.e.
\begin{equation}
\xi _t^{{\rm{mod}}} = D({f_\theta }({\bf{x}}_t),{\bf{GT}}_t^{{\rm{}}}),
\label{eqn:mod}
\end{equation}
where $D$ denotes the metric to measure the differences of the pair of images and we follow \cite{li2023error} to use KL divergence as the metric, ${f_\theta }$ denotes the process of the core network, and ${\bf{GT}}_{t}$ is ground truth image at iteration $t$.

\textbf{Cumulative error} $\xi_t^{{\rm{cumu}}}$ measures the amount of error that are accumulated for sequentially running the core network from the first iteration $T$ to the current iteration $t$. The error comes from the modular error and the input cumulative error, i.e.
\begin{equation}
\xi _t^{{\rm{cumu}}} = \xi_t^{{\rm{mod}}} + {\mu_t}\xi_{t+1}^{{\rm{cumu}}},
\label{eqn:cumu}
\end{equation}
where $\mu_{t}$ specifies how much the input cumulative error $\xi_{t+1}^{{\rm{cumu}}}$ is propagated to $\xi_t^{{\rm{cumu}}}$. The input error is defined as 
\begin{equation}
\xi_{t+1}^{{\rm{cumu}}} = D({\bf{x}}_t^{{\rm{back}}},{\bf{x}}_t^{{\rm{}}}),
\label{eqn:input}
\end{equation}
where ${\bf{x}}_t^{{\rm{back}}}$ is the input data at iteration $t$ according to the backward process. ${\bf{x}}_t^{{\rm{back}}}$ is obtained by linking the computation of all the previous iterations, i.e. $
{\bf{x}}_t^{{\rm{back}}} \sim{ {p_\theta }({{\bf{x}}_T})\prod\limits_{i= T}^{t+1} {{p_\theta }({{\bf{x}}_{i - 1}}|{{\bf{x}}_i})}}$, where ${p_\theta}(x_{t-1} | x_{t})$ denotes the processing of the core network with the parameter $\theta$ shared across different iterations. As proved in \cite{li2023error}, we have $\mu_{t} \ge 1$. This means that the input cumulative error is at least fully spread to the next iteration, and the cumulative errors cannot be neglected.

\begin{figure}

\begin{minipage}[ht]{.99\linewidth}
\centering
\vspace{-0pt}
\subfloat{\label{}\includegraphics[width=0.22\linewidth, height= 0.22\linewidth]{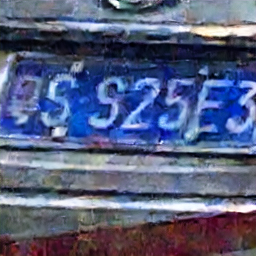}}\hspace{4pt}
\subfloat{\label{}\includegraphics[width=0.22\linewidth, height= 0.22\linewidth]{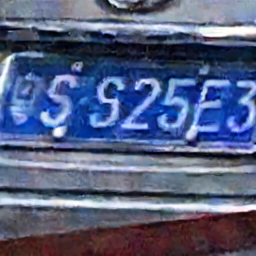}}\hspace{4pt}
\subfloat{\label{}\includegraphics[width=0.22\linewidth, height= 0.22\linewidth]{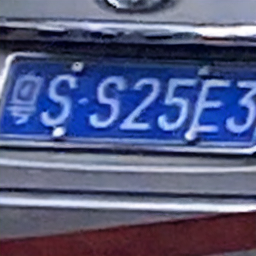}}\hspace{4pt}
\subfloat{\label{}\includegraphics[width=0.22\linewidth, height= 0.22\linewidth]{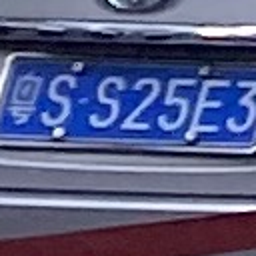}}\hspace{4pt}
\caption*{\scriptsize (a) Intermediate results of selected iterations in training of traditional DDM}
\end{minipage}

\begin{minipage}[ht]{.99\linewidth}
\centering
\vspace{-0pt}
\subfloat{\label{}\includegraphics[width=0.22\linewidth, height= 0.22\linewidth]{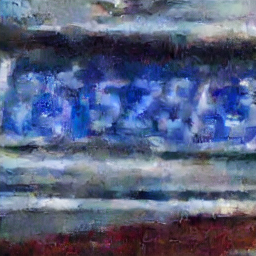}}\hspace{4pt}
\subfloat{\label{}\includegraphics[width=0.22\linewidth, height= 0.22\linewidth]{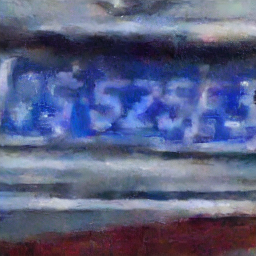}}\hspace{4pt}
\subfloat{\label{}\includegraphics[width=0.22\linewidth, height= 0.22\linewidth]{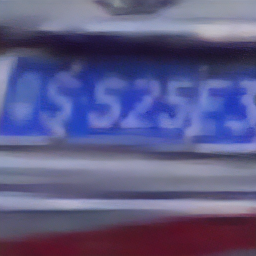}}\hspace{4pt}
\subfloat{\label{}\includegraphics[width=0.22\linewidth, height= 0.22\linewidth]{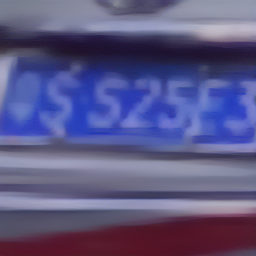}}\hspace{4pt}
\caption*{\scriptsize (b) Intermediate results of selected iterations in testing of traditional DDM}
\end{minipage}

\begin{minipage}[ht]{.99\linewidth}
\centering
\vspace{-0pt}
\subfloat{\label{}\includegraphics[width=0.22\linewidth, height= 0.22\linewidth]{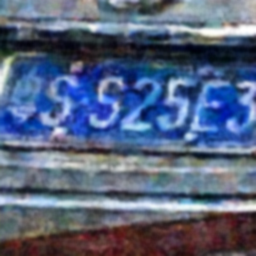}}\hspace{4pt}
\subfloat{\label{}\includegraphics[width=0.22\linewidth, height= 0.22\linewidth]{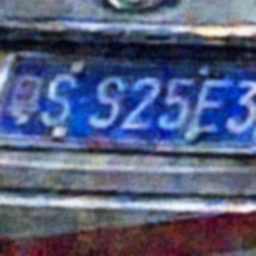}}\hspace{4pt}
\subfloat{\label{}\includegraphics[width=0.22\linewidth, height= 0.22\linewidth]{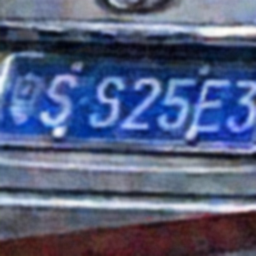}}\hspace{4pt}
\subfloat{\label{}\includegraphics[width=0.22\linewidth, height= 0.22\linewidth]{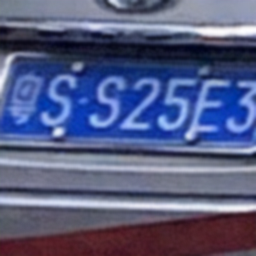}}\hspace{4pt}
\caption*{\scriptsize (c) Intermediate results of selected iterations in training of our consistent diffusion}
\end{minipage}

\begin{minipage}[ht]{.99\linewidth}
\centering
\vspace{-0pt}
\subfloat{\label{}\includegraphics[width=0.22\linewidth, height= 0.22\linewidth]{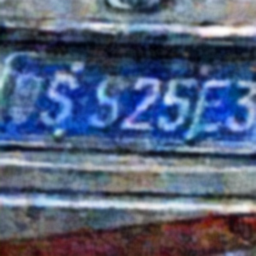}}\hspace{4pt}
\subfloat{\label{}\includegraphics[width=0.22\linewidth, height= 0.22\linewidth]{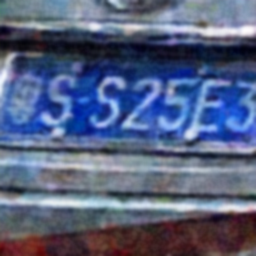}}\hspace{4pt}
\subfloat{\label{}\includegraphics[width=0.22\linewidth, height= 0.22\linewidth]{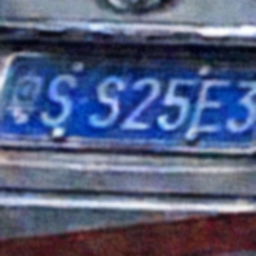}}\hspace{4pt}
\subfloat{\label{}\includegraphics[width=0.22\linewidth, height= 0.22\linewidth]{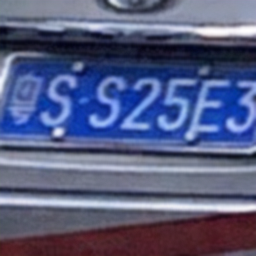}}\hspace{4pt}
\caption*{\scriptsize (d) Intermediate results of selected iterations in testing of our consistent diffusion}
\end{minipage}

\caption{ In traditional DDM, the gap of training and testing in (a) and (b) shows that although the modular error is minimized well during training, cumulative errors are not well controlled during testing. (c) and (d) show consistent accuracy between training and testing of our consistent diffusion, and demonstrate the effectiveness of our method to reduce cumulative errors during testing by optimizing them during training.
}
\label{fig:insight}
\end{figure}

\subsection{Traditional DDM}
In the training stage of traditional DDMs, at each iteration $t$, the input data ${\bf{x}}_t^{{\rm{train}}}$ is usually generated according to the forward processing, i.e. ${\bf{x}}_t^{{\rm{train}}}={\bf{x}}_t^{{\rm{forw}}}$.  ${\bf{x}}_t^{{\rm{forw}}}$ is obtained by the one-step distortion addition operation ${q}$, i.e.
${\bf{x}}_t^{{\rm{forw}}} \sim{q({{\bf{x}}_t}|{{\bf{x}}_0})}$, where ${{\bf{x}}_0}$ is the ground-truth image.
The loss function in the training stage of traditional DDMs is usually designed as
\begin{equation}
Loss =\sum_{t} \beta_t ({f_\theta }({\bf{x}}_t^{{\rm{train}}}),{\bf{GT}}_t^{{\rm{}}})=\sum_{t} \beta_t ({f_\theta }({\bf{x}}_t^{{\rm{forw}}}),{\bf{GT}}_t^{{\rm{}}}),
\label{eqn:loss_tradition}
\end{equation}
where $\beta_t$ is the metric at iteration $t$. This actually minimizes the modular error according to Eq. \ref{eqn:mod}.

However, in the testing stage where the backward process is performed, according to the cumulative error in Eq. \ref{eqn:cumu}, there exist two error sources, i.e. modular error $\xi _t^{{\rm{mod}}}$ and input cumulative error ${\mu _t}\xi _{t+1}^{{\rm{cumu}}}$. The loss function in Eq. \ref{eqn:loss_tradition} does not consider the cumulative error at all. During the training, the modular error can hardly be  minimized to zero at each iteration. Since $\mu_{t} \ge 1$, the cumulative error will increase through the series of iterations and cannot be neglected , as shown in Figs. \ref{fig:teasor} and \ref{fig:statistics}. Failing to consider the cumulative error in the training stage leads to uncontrolled errors into the testing stage and the accuracy gap between the training stage and the testing stage of traditional DDM, as shown in Figs. \ref{fig:teasor} and \ref{fig:statistics}.

\section{Method}
\subsection{Data-Consistent Training}

We notice that the limitation of the training method of traditional DDMs comes from the data inconsistency, i.e. the input data in the training stage is ${\bf{x}}_t^{{\rm{forw}}}$, while that in the testing stage is ${\bf{x}}_t^{{\rm{back}}}$.

At iteration $t$ of the training stage, we let the input change from ${\bf{x}}_t^{{\rm{forw}}}$ to ${\bf{x}}_t^{{\rm{back}}}$, i.e. ${\bf{x}}_t^{{\rm{train}}}={\bf{x}}_t^{{\rm{back}}}$, to avoid the data inconsistency. In this way, according to Eq. \ref{eqn:mod}, the new modular error changes to 
\begin{equation}
\hat{\xi} _t^{{\rm{mod}}} = D({f_\theta }({\bf{x}}_t^{{\rm{train}}}), {\bf{G}}{{\bf{T}}_t})= D({f_\theta }({\bf{x}}_t^{{\rm{back}}}), {\bf{G}}{{\bf{T}}_t}),
\end{equation}
and, according to Eq. \ref{eqn:input}, the input cumulative error changes to 
\begin{equation}
\hat{\xi} _{t+1}^{{\rm{cumu}}} = D({\bf{x}}_t^{{\rm{back}}},{\bf{x}}_t^{{\rm{train}}}) = D({\bf{x}}_t^{{\rm{back}}},{\bf{x}}_t^{{\rm{back}}}) = 0.
\end{equation}
Thus, according to Eq. \ref{eqn:cumu}, the cumulative error is
\begin{equation}
\hat{\xi} _t^{{\rm{cumu}}} = \hat{\xi} _t^{{\rm{mod}}} + {\mu _t} \cdot 0= \hat{\xi} _t^{{\rm{mod}}}.
\end{equation}
This indicates that by using ${\bf{x}}_t^{{\rm{back}}}$ as the input, the cumulative error at iteration $t$ is only decided by $\hat{\xi} _t^{{\rm{mod}}}$ and there does not exist any other error sources. Thus, we can follow traditional DDM to design the loss function to only measure the modular error, i.e. 
\begin{equation}
Loss = \sum_{t} \beta_t ({f_\theta }({\bf{x}}_t^{{\rm{train}}}),{\bf{GT}}_t^{{\rm{}}})= \sum_{t} \beta_t ({f_\theta }({\bf{x}}_t^{{\rm{back}}}),{\bf{GT}}_t^{{\rm{}}}),
\end{equation}
where $\beta_t$ is the metric at iteration $t$.

\subsection{Efficient Data-Consistent Training}

Although the proposed data-consistent training ensures no input cumulative error, it can lead to a increase in memory and computation costs in the training stage, because the input at each iteration $t$, i.e. ${\bf{x}}_t^{{\rm{back}}}$, needs to be generated by the series process of the core network from iteration $T$ to iteration $t+1$.  

For users that require low memory and computation costs in the training stage, we provide the alternative efficient version of our method, with the cost of small image quality drops. We introduce a two-step approach for generating the input at iteration $t$, named $\tilde{\bf{x}}_{t}^{\text{train}}$. First, we generate $\tilde{\bf{x}}_{t+1}^{\text{train}}$, i.e. the input data at iteration $t+1$, using the forward process. It is obtained by $\tilde{\bf{x}}_{t+1}^{\text{train}}={\bf{x}}_{t+1}^{\text{forw}}$, and ${\bf{x}}_{t+1}^{{\rm{forw}}} \sim{q({{\bf{x}}_{t+1}}|{{\bf{x}}_0})}$. Next, we use $f_{\theta}$ to obtain the processing result of iteration $t+1$, i.e. $ {f_\theta }({\tilde{\bf{x}}}_{t+1}^{{\rm{train}}})$. In this context, we have already introduced one-step cumulative error into $ {f_\theta }({\tilde{\bf{x}}}_{t+1}^{{\rm{train}}})$. The one-step error $\bf{e}$ is calculated by
\begin{equation}
{\bf{e}} = {f_\theta }({\tilde{\bf{x}}}_{t+1}^{{\rm{train}}}) - {\bf{GT}}_{t+1}.
\end{equation}
To simulate the cumulative errors during testing, we assume $\mu_t$ in Eq. \ref{eqn:cumu} is equal to 1 at each iteration. To ensure the strength of error is the same with that in consistent diffusion, we amplify the one-step error $\bf{e}$ by $\lambda$ times, and set $\lambda = T - t$ according to the number of iterations to obtain $\tilde{\bf{x}}_{t}^{\text{train}}$, i.e.
\begin{equation}
      \tilde{\bf{x}}_{t}^{\text{train}} = {f_\theta }({\tilde{\bf{x}}}_{t+1}^{{\rm{train}}}) + \lambda \cdot \bf{e},
\end{equation}
and the loss function changes to
\begin{equation}
Loss = \sum_{t} \beta_t ({f_\theta }({\tilde{\bf{x}}}_t^{{\rm{train}}}),{\bf{GT}}_t^{{\rm{}}}).
\end{equation}

\begin{table}[t]
\centering
\resizebox{0.5\textwidth}{!}{ 
\begin{tabular}{c|ccccc}
    \toprule
    Method & PSNR ($\uparrow$) & SSIM ($\uparrow$) & LPIPS ($\downarrow$) & FID ($\downarrow$) & CLIPIQA ($\uparrow$) \\
    \midrule
    DA-CLIP \cite{luo2024photo} & 29.67 & 0.8272 & 0.1262 & 128.28 & 0.1075 \\
    SwinIR \cite{liang2021swinir} & 29.82 & 0.8711 & 0.2293 & 197.59 & \underline{0.1973} \\
    AirNet \cite{li2022all} & 28.88 & 0.8702 & 0.2108 & 141.13 & 0.1129 \\
    DiffPlugin \cite{liu2024diff} & 23.65 & 0.7901 & 0.1442 & 141.75 & 0.0826 \\
    DiffUIR \cite{zheng2024selective} & 24.69 & 0.8166 & 0.1863 & \underline{76.03} & 0.0766 \\
    PromptIR \cite{potlapalli2306promptir} & 29.39 & \underline{0.8810} & 0.1291 & 166.38 & 0.1206 \\
    Restormer \cite{zamir2022restormer} & 30.02 & 0.8316 & \underline{0.1030} & 115.24 & 0.1249 \\
    MAXIM \cite{tu2022maxim} & \underline{30.81} & \textbf{0.9030} & 0.1362 & 195.92 & 0.1498 \\
    
    \midrule
    Blind2Unblind \cite{wang2022blind2unblind} & 25.32 & 0.7366 & 0.2189 & 179.99 & 0.1018 \\
    KBNet \cite{zhang2023kbnet} & 30.04 & 0.8753 & 0.2546 & 128.43 & \underline{0.2073} \\
    Pretrained-IPT \cite{chen2021pre} & 29.71 & 0.8178 & 0.2874 & 208.46 & 0.1603 \\
    SADNet \cite{sun2022sadnet} & 27.92 & 0.8332 & \underline{0.1159} & 116.97 & 0.1186 \\
    \midrule
    Ours (efficient) & \underline{30.42} & 0.8784 & 0.1589 & \underline{63.97} & 0.1620 \\
    Ours & \textbf{32.25} & \underline{0.8989} & \textbf{0.0812} & \textbf{61.82} & \textbf{0.2090} \\
    \bottomrule 
\end{tabular}
}
\caption{Quantitative comparison of denoising on Urban100 \cite{huang2015single}. We compare 8 general and 4 task-specific restoration models.}
\label{table:denoise}
\end{table}

\begin{table}[t]
\centering
\resizebox{0.5\textwidth}{!}{ 
\begin{tabular}{c|ccccc}
    \toprule
    Method & PSNR ($\uparrow$) & SSIM ($\uparrow$) & LPIPS ($\downarrow$) & FID ($\downarrow$) & CLIPIQA ($\uparrow$) \\
    \midrule
    DA-CLIP \cite{luo2024photo} & \underline{33.91} & \underline{0.9261} & \underline{0.0942} & \textbf{48.28} & 0.1239 \\
    SwinIR \cite{liang2021swinir}  & 25.22 & 0.8369 & 0.2621 & 123.28 & 0.0791 \\
    AirNet \cite{li2022all} & 19.34 & 0.6734 & 0.3336 & 221.33 & 0.0878 \\
    DiffPlugin \cite{liu2024diff}  & 22.71 & 0.7371 & 0.2002 & 188.15 & 0.1066 \\
    DiffUIR \cite{zheng2024selective} & 19.66 & 0.6742 & 0.3019 & 217.10 & 0.0858 \\
    PromptIR \cite{potlapalli2306promptir} & 20.76 & 0.6828 & 0.2792 & 190.93 & 0.0963 \\
    Restormer \cite{zamir2022restormer} & \underline{31.46} & 0.9043 & 0.1289 & 111.03 & 0.1297 \\
     MAXIM \cite{tu2022maxim} & 28.25 & 0.8963 & 0.1397 & 112.48 & 0.1258 \\
     \midrule
    WGWS \cite{zhu2023learning} & 20.15 & 0.6678 & 0.2901 & 285.95 & 0.0648 \\
    KBNet \cite{zhang2023kbnet} & 29.13 & 0.8991 & 0.1362 & 114.43 & \underline{0.1335} \\
    Pretrained-IPT \cite{chen2021pre} & 20.28 & 0.6925 & 0.3385 & 224.95 & 0.1193 \\
    \midrule
    Ours (efficient)& 31.08 & \underline{0.9219} & \underline{0.1117} & \underline{56.63}& \underline{0.1510} \\
    Ours & \textbf{34.22} & \textbf{0.9359} & \textbf{0.0747} & \underline{98.85} & \textbf{0.2276} \\
    \bottomrule 
    
\end{tabular}
}
\caption{Quantitative comparison of derain on Rain-100H \cite{jiang2020multi}. We compare 8 general and 3 task-specific restoration models.}
\label{table:derain}
\end{table}

\begin{table}[t]
\centering
\resizebox{0.5\textwidth}{!}{ 
\begin{tabular}{c|ccccc}
    \toprule
    Method & PSNR ($\uparrow$) & SSIM ($\uparrow$) & LPIPS ($\downarrow$) & FID ($\downarrow$) & CLIPIQA ($\uparrow$) \\
    \midrule
    DA-CLIP \cite{luo2024photo} & \underline{30.42} & \underline{0.9668} & \underline{0.0341} & \textbf{6.76} & 0.1373 \\
    SwinIR \cite{liang2021swinir}  & 25.11 & 0.8314 & 0.1282 & 44.89 & 0.1381 \\
    AirNet \cite{li2022all} & 22.39 & 0.8630 & 0.1363 & 30.45 & 0.1457 \\
    DiffPlugin \cite{liu2024diff}  & 24.60 & 0.8752 & 0.1069 & 32.26 & 0.1291 \\
    DiffUIR \cite{zheng2024selective} & 25.76 & 0.8609 & 0.1579 & 68.72 & \textbf{0.1889} \\
    PromptIR \cite{potlapalli2306promptir} & 26.79 & 0.8214 & 0.1307 & 23.39 & 0.1447 \\
    Restormer \cite{zamir2022restormer} & 28.81 & 0.9537 & 0.0460 & 9.16 & 0.1328 \\
     MAXIM \cite{tu2022maxim} & 26.04 & 0.9360 & 0.0456 & \underline{7.21} & 0.1591 \\
    \midrule
    mixDehazeNet \cite{lu2024mixdehazenet} & 29.13 & 0.9547 & 0.0513 & 46.53 & 0.1688 \\
    dehazeFormer \cite{song2023vision} & \underline{30.63} & \underline{0.9741} & \underline{0.0379} & 9.21 & 0.1386 \\
    WGWS \cite{zhu2023learning} & 21.78 & 0.8472 & 0.1645 & 93.20 & \underline{0.1698} \\
    \midrule
    Ours (efficient)& 29.19 & 0.9309 & 0.1156 & 16.68 & 0.1604 \\
    Ours & \textbf{31.12} & \textbf{0.9747} & \textbf{0.0321} & \underline{7.43} & \underline{0.1860} \\
    \bottomrule 
\end{tabular}
}
\caption{Quantitative comparison of dehaze on RESIDE-6k \cite{li2018benchmarking}. We compare 8 general and 3 task-specific restoration models.}
\label{table:dehaze}
\end{table}

\begin{table}[t]
\centering
\resizebox{0.5\textwidth}{!}{ 
\begin{tabular}{c|ccccc}
    \toprule
    Method & PSNR ($\uparrow$) & SSIM ($\uparrow$) & LPIPS ($\downarrow$) & FID ($\downarrow$) & CLIPIQA ($\uparrow$) \\
    \midrule
    SwinIR \cite{liang2021swinir}  & 25.99 & 0.7657 & 0.1164 & 25.04 & 0.0733 \\
    ZEDUSR \cite{xu2023zero} & 30.28 & \underline{0.8976} & 0.0334 & \underline{10.22} & \underline{0.0864} \\
    C$^{2}$-Matching \cite{jiang2022reference} & 31.79 & 0.8264 & 0.0417 & 14.75 & 0.0773 \\
    DCSR \cite{wang2021dual} & \underline{34.25} & \underline{0.8785} & \underline{0.0263} & \underline{8.78} & 0.0839 \\
    DZSR \cite{zhang2022self} & 31.39 & 0.8602 & 0.0820 & 18.87 & 0.1042 \\
    EDSR \cite{yue2024kedusr} & 32.35 & 0.6543 & 0.0295 & 17.57 & 0.0778 \\
    MASA \cite{lu2021masa} & 29.94 & 0.8041 & 0.0763 & 18.63 & \textbf{0.0892} \\
    RCAN \cite{xu2023zero} & 32.45 & 0.8555 & \underline{0.0293} & 18.46 & 0.0783 \\
    Real-ESRGAN \cite{wang2021real} & 27.80 & 0.7660 & 0.1637 & 28.91 & \underline{0.0889} \\
    SRGAN \cite{ledig2017photo} & 29.27 & 0.7622 & 0.0935 & 14.96 & 0.0852 \\
    SRNTT \cite{zhang2019image} & 31.42 & 0.8224 & 0.0449 & 17.76 & 0.0756 \\
    TTSR \cite{yang2020learning} & 29.13 & 0.7767 & 0.0498 & 14.75 & 0.0874 \\
    Ours (efficient)& \underline{33.78} & 0.8760 & 0.0535 & 18.59 & 0.0873 \\
 Ours & \textbf{35.16} & \textbf{0.9047} & \textbf{0.0158} & \textbf{7.43} & 0.0860 \\
    \bottomrule 
\end{tabular}
} 
\caption{Quantitative comparison of dual camera super-resolution on CameraFusion \cite{wang2021dual}. We compare 12 task-specific methods.}
\label{table:dual}
\end{table}

\begin{table*}[t]
\centering
\resizebox{0.9\textwidth}{!}{ 
\begin{tabular}{c|ccccc|ccccc}
    \toprule
    \multirow{2}{*}{Method} & \multicolumn{5}{c|}{\textbf{RealSR }} & \multicolumn{5}{c}{\textbf{CameraFusion}} \\
    \cline{2-11}
    & PSNR ($\uparrow$) & SSIM ($\uparrow$) & LPIPS ($\downarrow$) & FID ($\downarrow$) & CLIPIQA ($\uparrow$) & PSNR ($\uparrow$) & SSIM ($\uparrow$) & LPIPS ($\downarrow$) & FID ($\downarrow$) & CLIPIQA ($\uparrow$) \\
    \midrule
    DA-CLIP \cite{luo2024photo}  & 25.24 & 0.7532 & 0.3639 & 116.22 & 0.2054 & 21.54 & 0.7554 & 0.1450 & 42.34 & 0.0873 \\
    SwinIR \cite{liang2021swinir}  & 24.58 & 0.7477 & 0.3605 & 102.18 & 0.1934 & 26.98 & 0.7473 & 0.1597 & 35.21 & 0.0807 \\
    AirNet \cite{li2022all} & \textbf{26.03} & \underline{0.7611} & 0.3556 & 111.35 & 0.1993 & 26.15 & 0.7974 & 0.1539 & 35.72 & 0.0889 \\
    DiffPlugin \cite{liu2024diff}  & 24.59 & 0.7431 & 0.3797 & 102.10 & 0.2003 & 29.86 & 0.7901 & 0.1442 & 31.75 & 0.0862 \\
    DiffUIR \cite{zheng2024selective} & 25.23 & \underline{0.7612} & 0.3681 & 129.48 & \underline{0.2076} & 29.40 & 0.7966 & 0.0621 & 46.03 & 0.0764 \\
    PromptIR \cite{potlapalli2306promptir} & 20.47 & 0.7437 & \textbf{0.3327} & 149.86 & \textbf{0.2223} & 26.08 & 0.6915 & 0.1286 & 31.03 & 0.0890 \\
    Restormer \cite{zamir2022restormer} & \underline{25.91} & 0.7315 & 0.3533 & 116.48 & \textbf{0.2060} & 30.35 & \underline{0.8365} & \textbf{0.0269} & 29.23 & 0.0799 \\
    MAXIM \cite{tu2022maxim} & 21.26 & 0.7415 & 0.3660 & 105.86 & 0.2059 & 22.72 & 0.7490 & 0.1742 & 89.54 & 0.0493 \\
    \bottomrule
    Real-ESRGAN \cite{wang2021real} & 24.27 & 0.7432 & 0.3658 & 104.87 & 0.1953 & 27.48 & 0.7619 & 0.1575 & 40.66 & 0.0921 \\
    DiffIR \cite{xia2023diffir} & 25.68 & \textbf{0.7772} & 0.3461 & 105.48 & 0.1892 & \underline{30.36} & 0.8051 & 0.1382 & 34.08 & 0.0815 \\
    HAT \cite{chen2023hat} & 25.86 & 0.7552 & 0.3581 & 110.26 & 0.2001 & 29.40 & 0.7966 & \underline{0.0601} & 45.93 & 0.0765 \\
    ResShift \cite{yue2024resshift} & 24.59 & 0.7431 & 0.3797 & 102.10 & 0.1903 & 27.28 & 0.7824 & 0.1332 & \underline{28.58} & 0.0765 \\
    SeeSR \cite{wu2024seesr} & 24.79 & 0.7246 & 0.3597 & \textbf{79.64} & 0.1912 & 28.42 & 0.7674 & 0.1488 & \underline{28.31} & 0.0891 \\
    StableSR \cite{wang2024exploiting} & 25.81 & 0.7401 & 0.4586 & \underline{84.53} & 0.1961 & \underline{30.71} & \underline{0.8268} & 0.1561 & 44.29 & \textbf{0.1181} \\
    SRFormer \cite{zhou2023srformer} & 24.87 & 0.7562 & 0.3567 & 96.19 & 0.1712 & 28.60 & 0.7871 & 0.1511 & 35.54 & 0.0764 \\
    \bottomrule
    Ours (efficient)  & 25.09 & 0.7430 & \underline{0.3453} & \underline{89.52} & 0.1899 & 28.68 & 0.7965 & 0.1767 & 59.85 & \underline{0.1123}\\
    Ours & \underline{25.96} & 0.7471 & \underline{0.3402} & 99.02 & 0.1957 & \textbf{31.41} & \textbf{0.8391} & \underline{0.0427} & \textbf{27.64} & \underline{0.1006} \\
    \bottomrule
\end{tabular}
}
\caption{Quantitative comparison of single image super-resolution on RealSR \cite{ji2020real} and CameraFusion \cite{wang2021dual}.}
\label{table:sisr}
\end{table*}

\begin{table}[t]
\centering
\resizebox{0.5\textwidth}{!}{ 
\begin{tabular}{c|c|ccccc}
    \toprule
    Method & Dataset & PSNR (\(\uparrow\)) & SSIM (\(\uparrow\)) & LPIPS (\(\downarrow\)) & FID (\(\downarrow\)) & CLIPIQA (\(\uparrow\)) \\
    \midrule
    \textbf{ResShift} \cite{yue2024resshift} & RealSR & 24.59 & 0.7431 & \underline{0.3797} & 102.10 & \underline{0.1903}\\
    \textbf{LDM} \cite{rombach2022high} & RealSR & \underline{25.19} & \underline{0.7586} & \underline{0.3691} & 111.88 & 0.1208 \\
    \textbf{\textbf{$L_{nll}+L_{reg}$}} \cite{li2023error} & RealSR & 24.71 & 0.7289 & 0.4072 & \textbf{75.89} & 0.0996 \\
     \textbf{LDM+DCT} & RealSR & \textbf{27.60} & \textbf{0.7821} & 0.3909 & \underline{86.49} & \underline{0.1905}  \\
    \textbf{ResShift+DCT (Ours)} & RealSR & \underline{25.96} & \underline{0.7471} & \textbf{0.3402} & \underline{99.02} & \textbf{0.1957}  \\
    \bottomrule
    \textbf{ResShift} \cite{yue2024resshift} & CameraFusion & 27.28 & \underline{0.7824} & \underline{0.1332} & \underline{28.58} & 0.0765 \\
    \textbf{LDM} \cite{rombach2022high} & CameraFusion & 26.15 & 0.7465 & 0.2211 & \underline{44.13} & \textbf{0.1317} \\
    \textbf{\textbf{$L_{nll}+L_{reg}$}} \cite{li2023error} & CameraFusion & \underline{30.23} & 0.7425 & \underline{0.0821} & 56.90 & 0.0859 \\
    \textbf{LDM+DCT} & CameraFusion & \underline{28.68} & \underline{0.7965} & 0.1767 & 59.85 & \underline{0.1123}\\
    \textbf{ResShift+DCT (Ours)} & CameraFusion & \textbf{31.41} & \textbf{0.8391} & \textbf{0.0427} & \textbf{27.64} & \underline{0.1006}  \\
    \bottomrule
\end{tabular}
} 
\caption{Ablation study about our data-consistent training. `DCT' is short for the proposed data-consistent training.}
\label{table:other_ablation}
\end{table}

\captionsetup[subfloat]{labelsep=none,format=plain,labelformat=empty}
\begin{figure*}
\vspace{-20pt}
\begin{minipage}[ht]{.99\linewidth}
\centering
\subfloat{\label{}\includegraphics[width=0.11\linewidth]{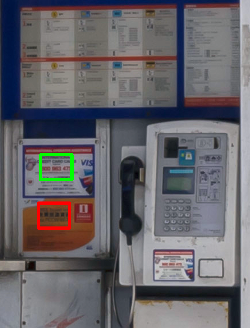}}\hspace{4pt}
\subfloat{\label{}\includegraphics[width=0.11\linewidth]{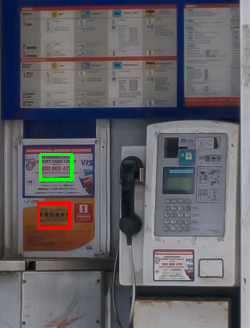}}\hspace{4pt}
\subfloat{\label{}\includegraphics[width=0.11\linewidth]{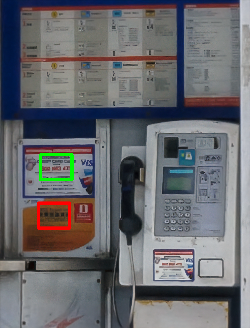}}\hspace{4pt}
\subfloat{\label{}\includegraphics[width=0.11\linewidth]{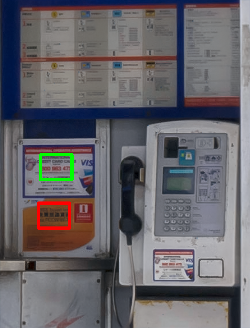}}\hspace{4pt}
\subfloat{\label{}\includegraphics[width=0.11\linewidth]{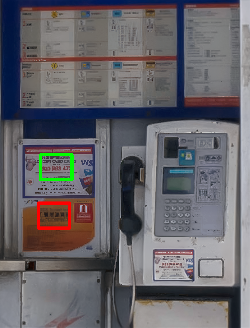}}\hspace{4pt}
\subfloat{\label{}\includegraphics[width=0.11\linewidth]{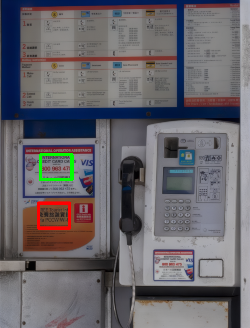}}\hspace{4pt}
\subfloat{\label{}\includegraphics[width=0.11\linewidth]{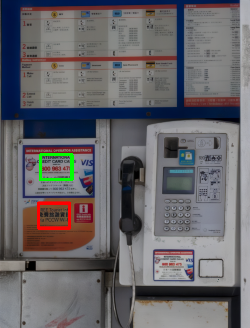}}\hspace{4pt}
\subfloat{\label{}\includegraphics[width=0.11\linewidth]{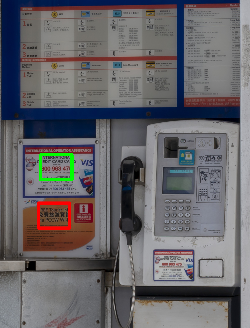}}\hspace{4pt}
\end{minipage}
\begin{minipage}[ht]{.99\linewidth}
\centering
\subfloat{\label{}\includegraphics[width=0.11\linewidth]{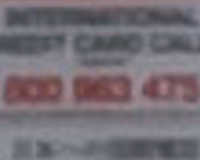}}\hspace{4pt}
\subfloat{\label{}\includegraphics[width=0.11\linewidth]{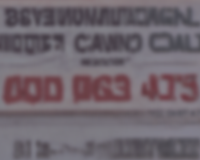}}\hspace{4pt}
\subfloat{\label{}\includegraphics[width=0.11\linewidth]{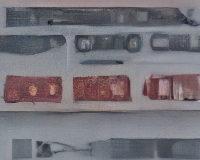}}\hspace{4pt}
\subfloat{\label{}\includegraphics[width=0.11\linewidth]{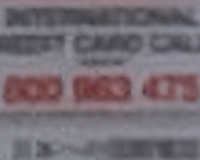}}\hspace{4pt}
\subfloat{\label{}\includegraphics[width=0.11\linewidth]{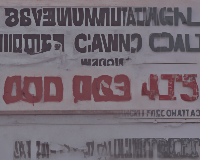}}\hspace{4pt}
\subfloat{\label{}\includegraphics[width=0.11\linewidth]{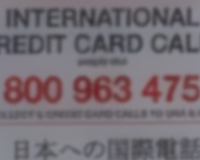}}\hspace{4pt}
\subfloat{\label{}\includegraphics[width=0.11\linewidth]{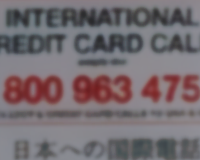}}\hspace{4pt}
\subfloat{\label{}\includegraphics[width=0.11\linewidth]{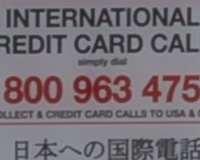}}\hspace{4pt}
\end{minipage}
\begin{minipage}[ht]{.99\linewidth}
\centering
\subfloat[Input]{\label{}\includegraphics[width=0.11\linewidth]{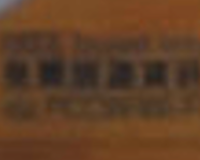}}\hspace{4pt}
\subfloat[MAXIM]{\label{}\includegraphics[width=0.11\linewidth]{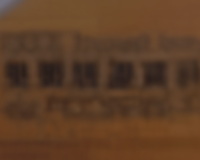}}\hspace{4pt}
\subfloat[ResShift]{\label{}\includegraphics[width=0.11\linewidth]{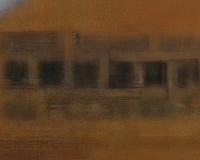}}\hspace{4pt}
\subfloat[AirNet]{\label{}\includegraphics[width=0.11\linewidth]{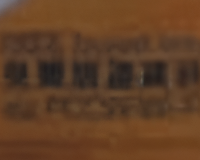}}\hspace{4pt}
\subfloat[SeeSR]{\label{}\includegraphics[width=0.11\linewidth]{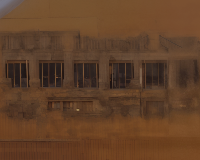}}\hspace{4pt}
\subfloat[Ours (efficient)]{\label{}\includegraphics[width=0.11\linewidth]{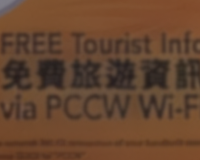}}\hspace{4pt}
\subfloat[Ours]{\label{}\includegraphics[width=0.11\linewidth]{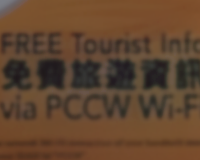}}\hspace{4pt}
\subfloat[GT]{\label{}\includegraphics[width=0.11\linewidth]{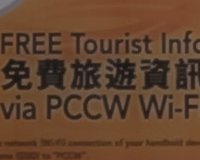}}\hspace{4pt}
\end{minipage}
\vspace{-5pt}
\caption{Example results in single image super-resolution on RealSR. The marked regions are enlarged. Due to page limits, we show results of the top 4 comparison methods here. Complete results and additional comparisons are available in the supplementary materials.}
\label{fig:sisr1}
\end{figure*}

\captionsetup[subfloat]{labelsep=none,format=plain,labelformat=empty}
\begin{figure*}
\begin{minipage}[ht]{.99\linewidth}
\centering
\subfloat{\label{}\includegraphics[width=0.11\linewidth]{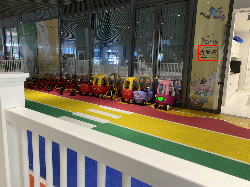}}\hspace{4pt}
\subfloat{\label{}\includegraphics[width=0.11\linewidth]{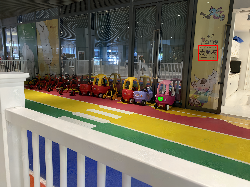}}\hspace{4pt}
\subfloat{\label{}\includegraphics[width=0.11\linewidth]{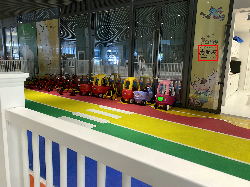}}\hspace{4pt}
\subfloat{\label{}\includegraphics[width=0.11\linewidth]{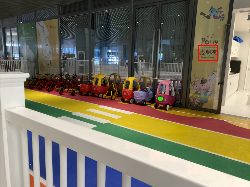}}\hspace{4pt}
\subfloat{\label{}\includegraphics[width=0.11\linewidth]{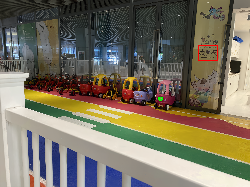}}\hspace{4pt}
\subfloat{\label{}\includegraphics[width=0.11\linewidth]{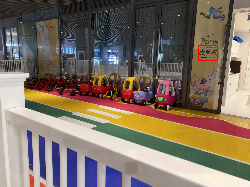}}\hspace{4pt}
\subfloat{\label{}\includegraphics[width=0.11\linewidth]{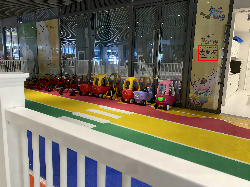}}\hspace{4pt}
\subfloat{\label{}\includegraphics[width=0.11\linewidth]{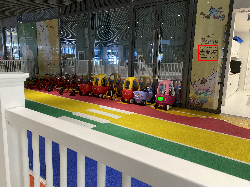}}\hspace{4pt}
\end{minipage}
\begin{minipage}[ht]{.99\linewidth}
\centering
\subfloat{\label{}\includegraphics[width=0.11\linewidth]{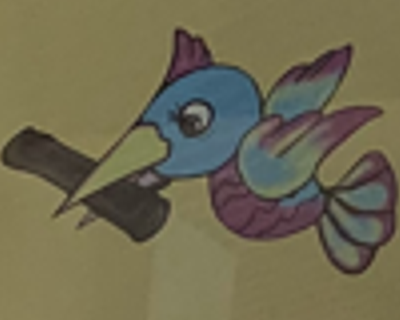}}\hspace{4pt}
\subfloat{\label{}\includegraphics[width=0.11\linewidth]{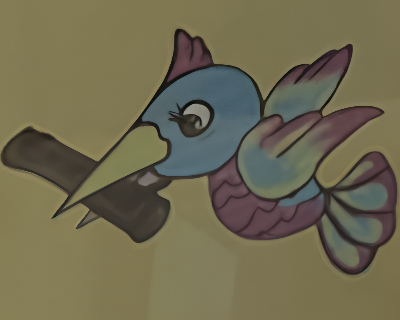}}\hspace{4pt}
\subfloat{\label{}\includegraphics[width=0.11\linewidth]{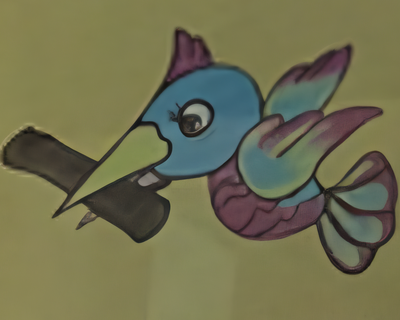}}\hspace{4pt}
\subfloat{\label{}\includegraphics[width=0.11\linewidth]{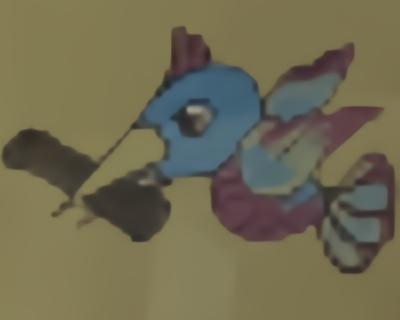}}\hspace{4pt}
\subfloat{\label{}\includegraphics[width=0.11\linewidth]{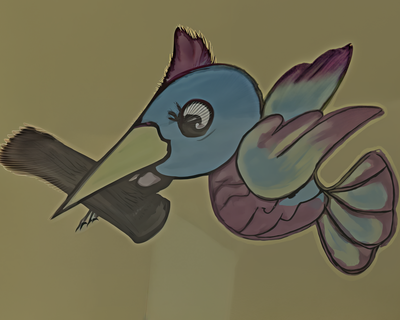}}\hspace{4pt}
\subfloat{\label{}\includegraphics[width=0.11\linewidth]{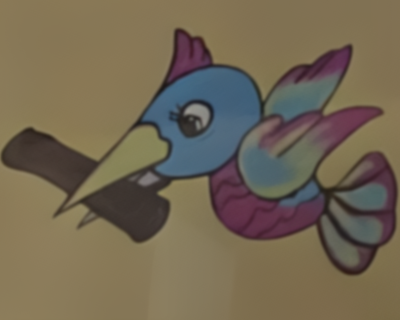}}\hspace{4pt}
\subfloat{\label{}\includegraphics[width=0.11\linewidth]{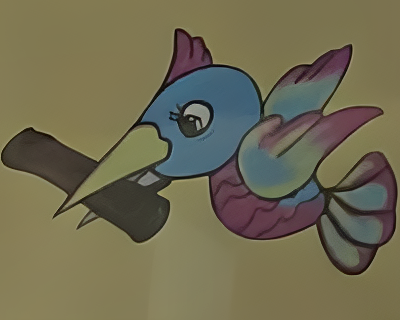}}\hspace{4pt}
\subfloat{\label{}\includegraphics[width=0.11\linewidth]{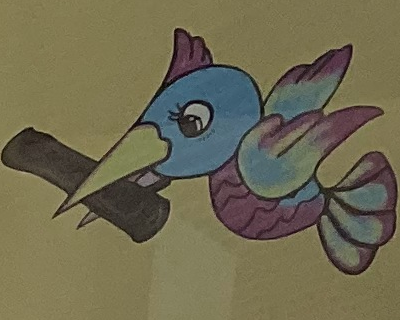}}\hspace{4pt}
\end{minipage}
\begin{minipage}[ht]{.99\linewidth}
\centering
\subfloat[Input]{\label{}\includegraphics[width=0.11\linewidth]{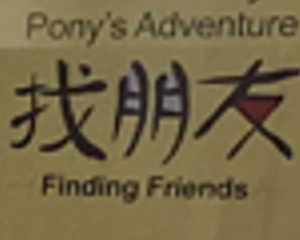}}\hspace{4pt}
\subfloat[MAXIM]{\label{}\includegraphics[width=0.11\linewidth]{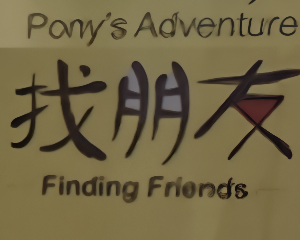}}\hspace{4pt}
\subfloat[ResShift]{\label{}\includegraphics[width=0.11\linewidth]{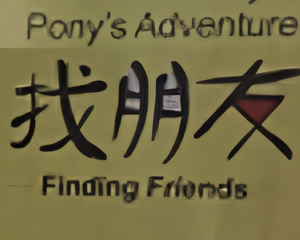}}\hspace{4pt}
\subfloat[AirNet]{\label{}\includegraphics[width=0.11\linewidth]{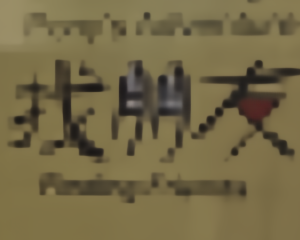}}\hspace{4pt}
\subfloat[SeeSR]{\label{}\includegraphics[width=0.11\linewidth]{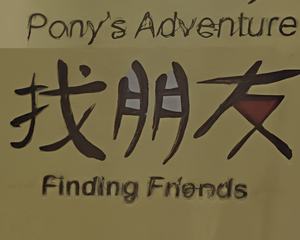}}\hspace{4pt}
\subfloat[Ours (efficient)]{\label{}\includegraphics[width=0.11\linewidth]{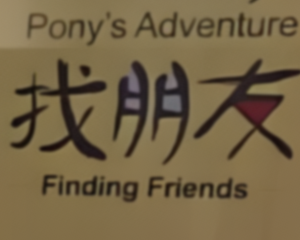}}\hspace{4pt}
\subfloat[Ours]{\label{}\includegraphics[width=0.11\linewidth]{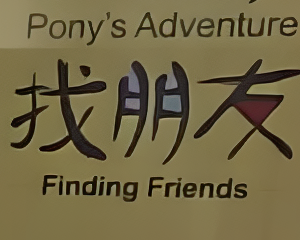}}\hspace{4pt}
\subfloat[GT]{\label{}\includegraphics[width=0.11\linewidth]{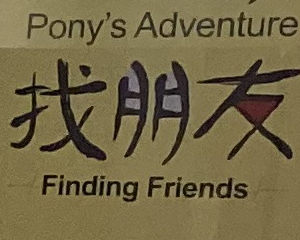}}\hspace{4pt}
\end{minipage}
\vspace{-5pt}
\caption{Example results in single image super-resolution on CameraFusion. Please see more results in the supplementary materials.}
\label{fig:sisr2}
\end{figure*}

\captionsetup[subfloat]{labelsep=none,format=plain,labelformat=empty}
\begin{figure*}
\vspace{-5pt}
\begin{minipage}[ht]{.99\linewidth}
\centering
\subfloat{\label{}\includegraphics[width=0.11\linewidth]{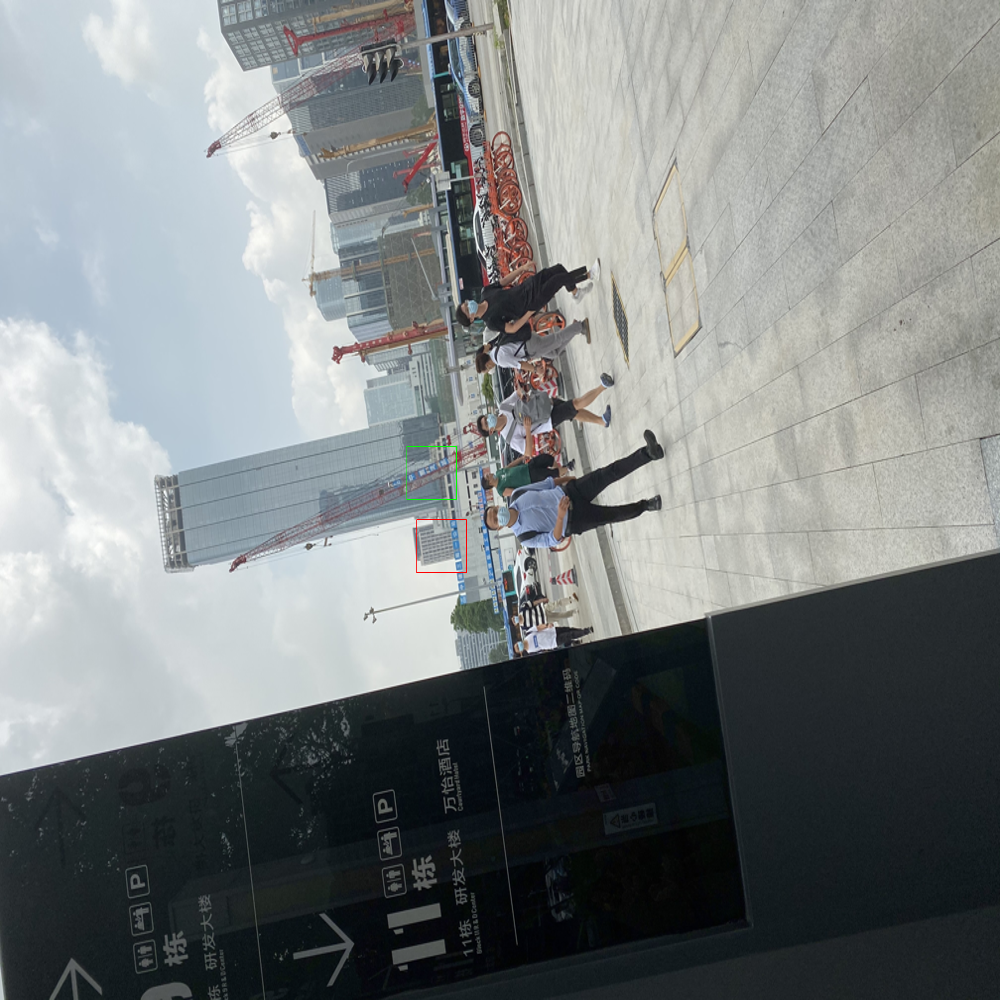}}\hspace{4pt}
\subfloat{\label{}\includegraphics[width=0.11\linewidth]{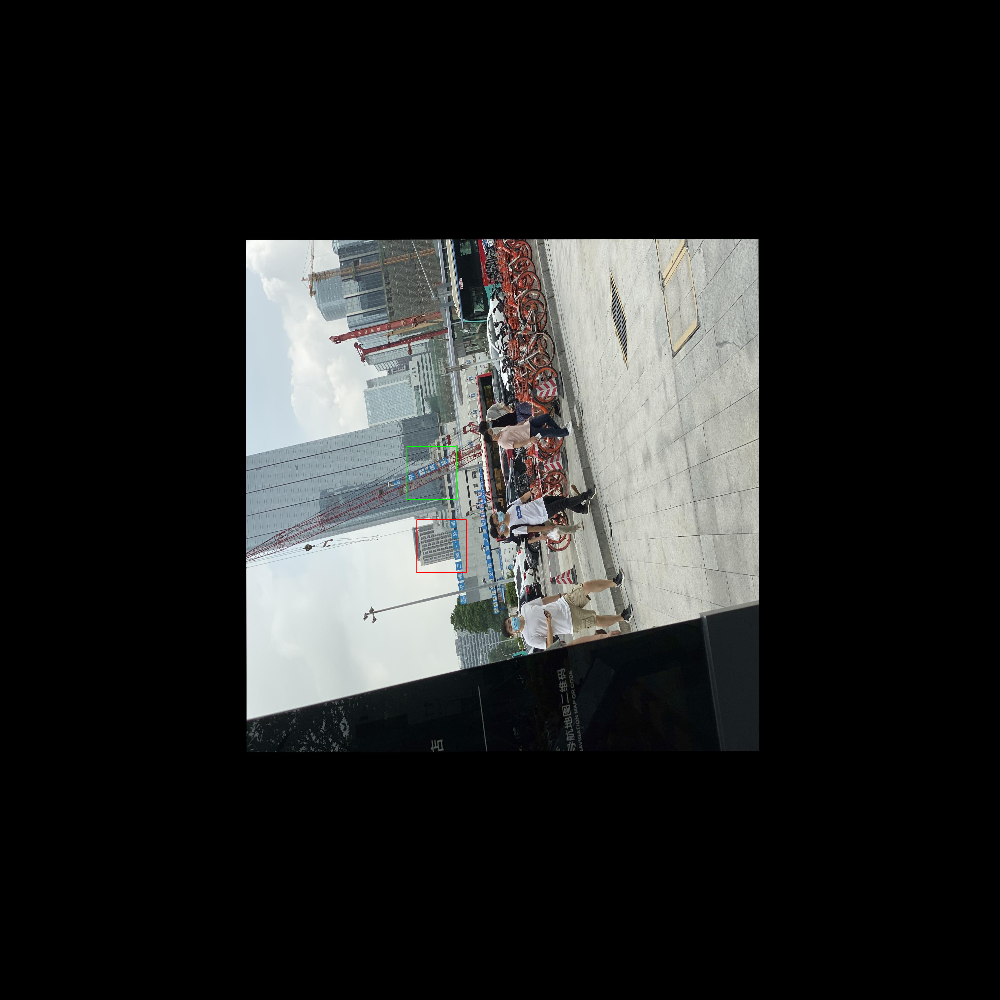}}\hspace{4pt}
\subfloat{\label{}\includegraphics[width=0.11\linewidth]{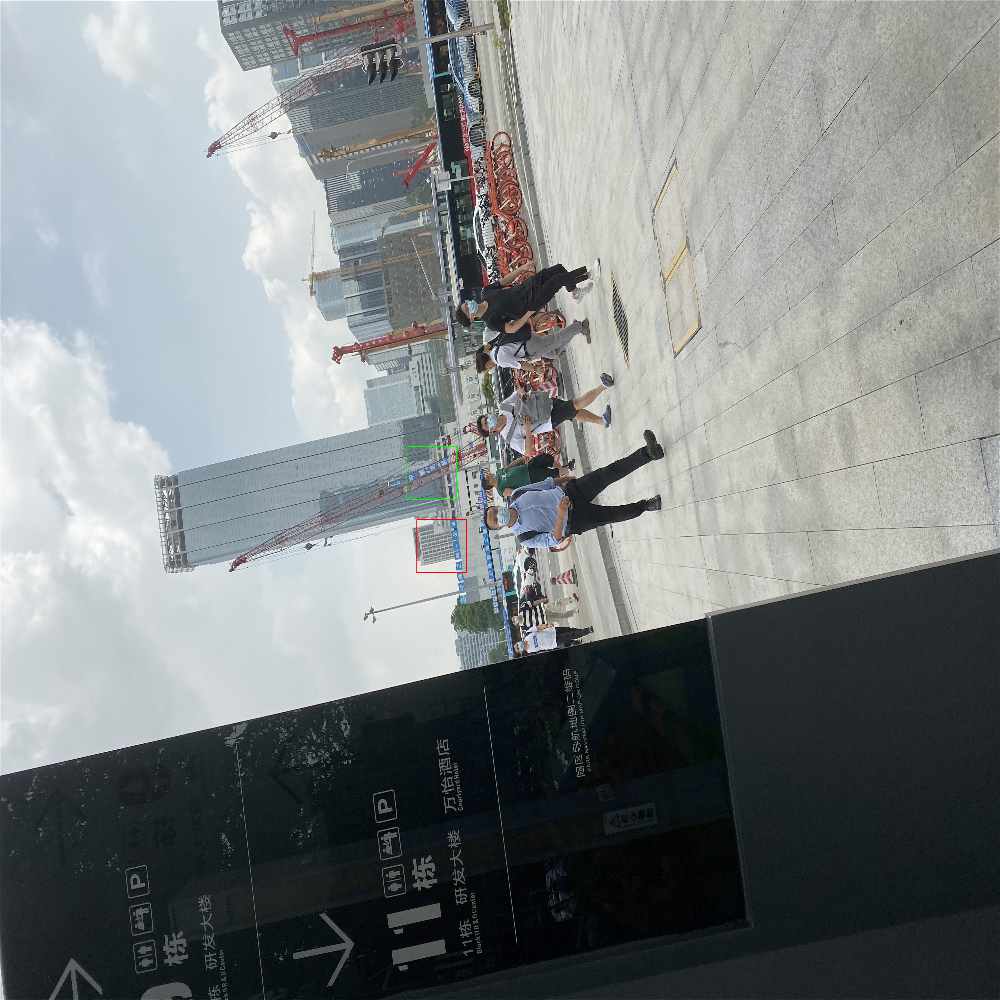}}\hspace{4pt}
\subfloat{\label{}\includegraphics[width=0.11\linewidth]{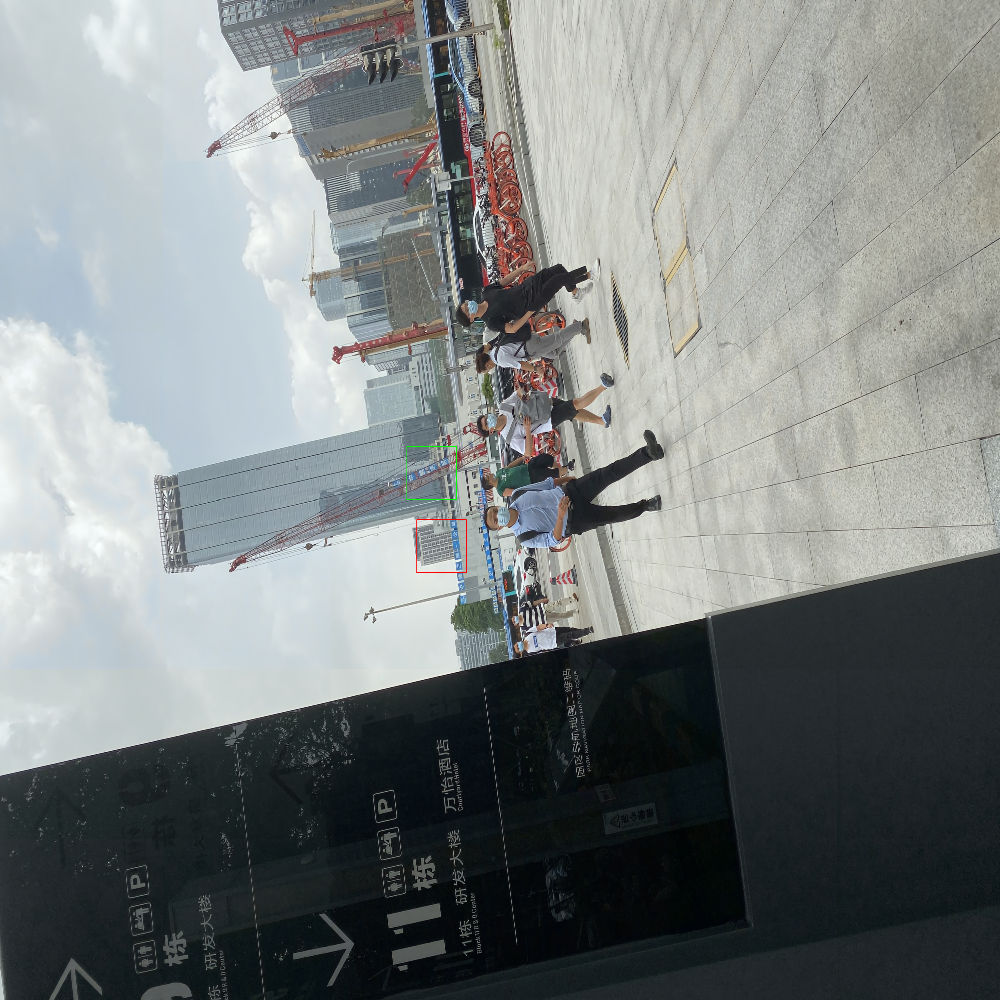}}\hspace{4pt}
\subfloat{\label{}\includegraphics[width=0.11\linewidth]{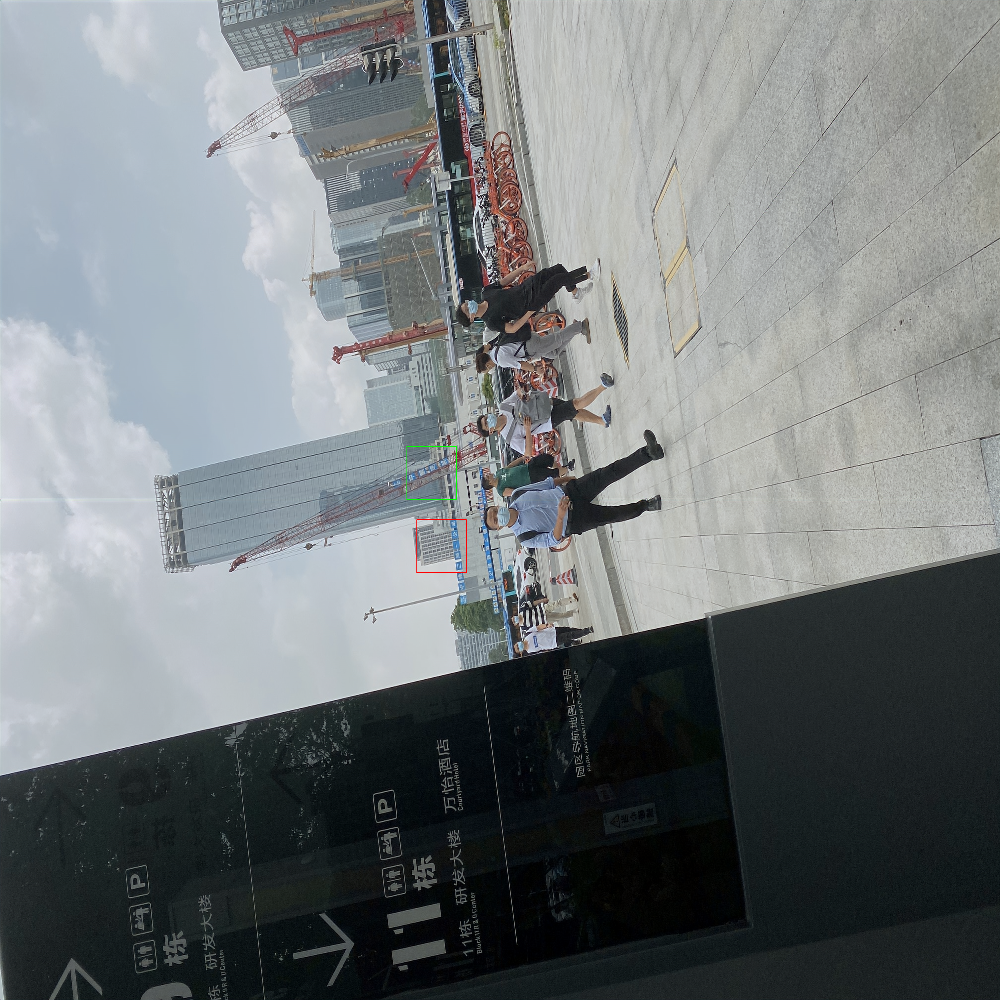}}\hspace{4pt}
\subfloat{\label{}\includegraphics[width=0.11\linewidth]{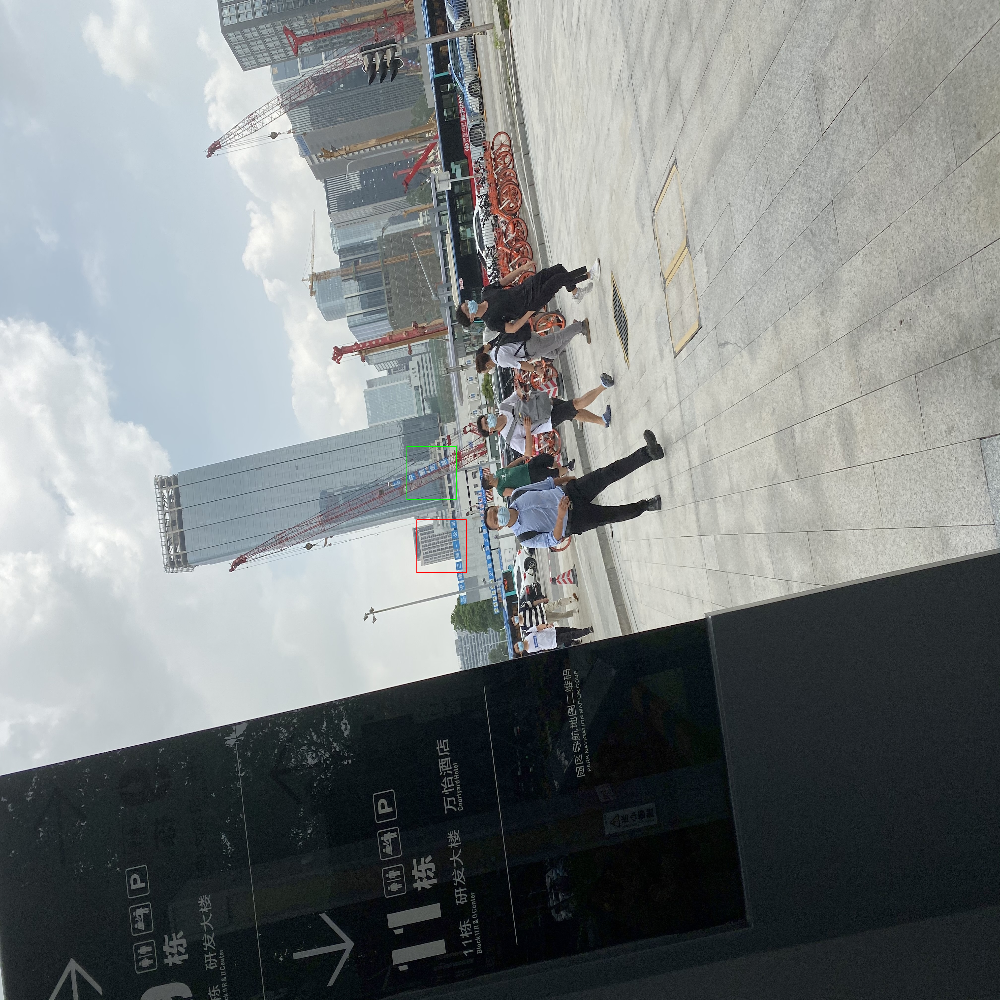}}\hspace{4pt}
\subfloat{\label{}\includegraphics[width=0.11\linewidth]{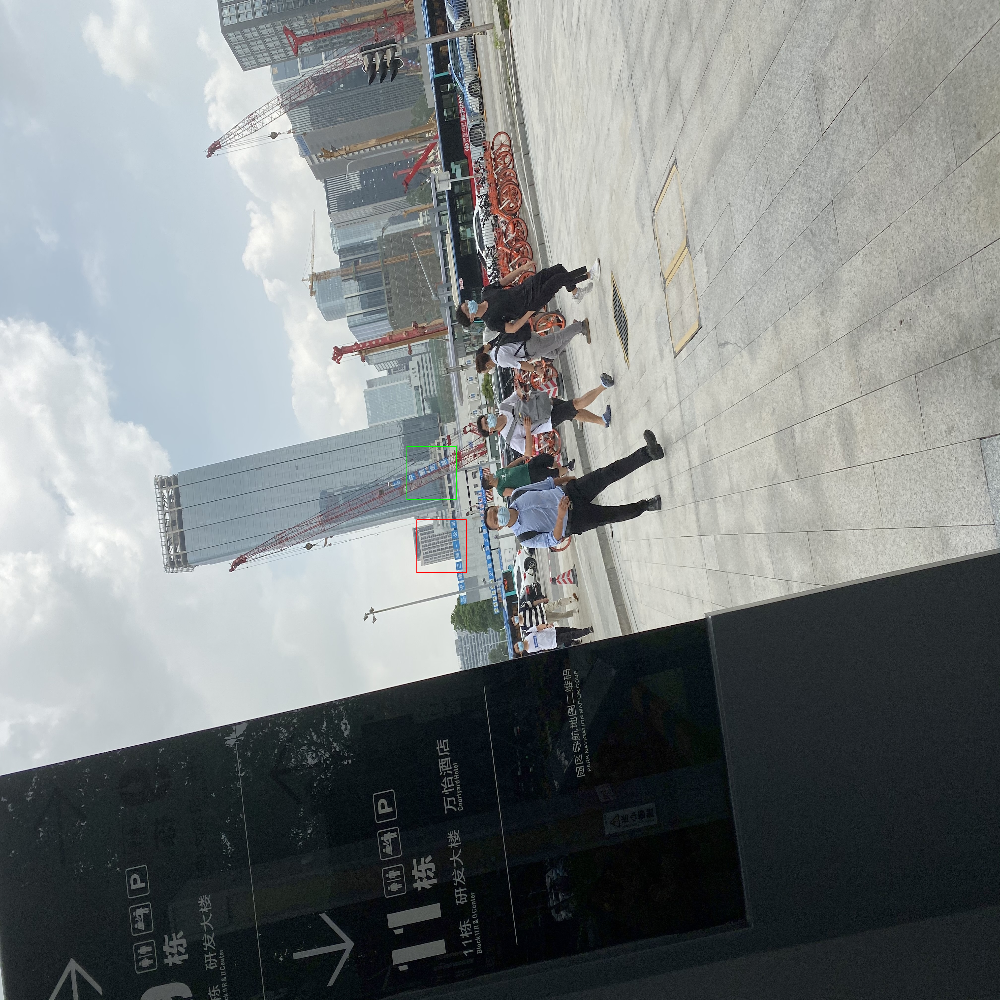}}\hspace{4pt}
\subfloat{\label{}\includegraphics[width=0.11\linewidth]{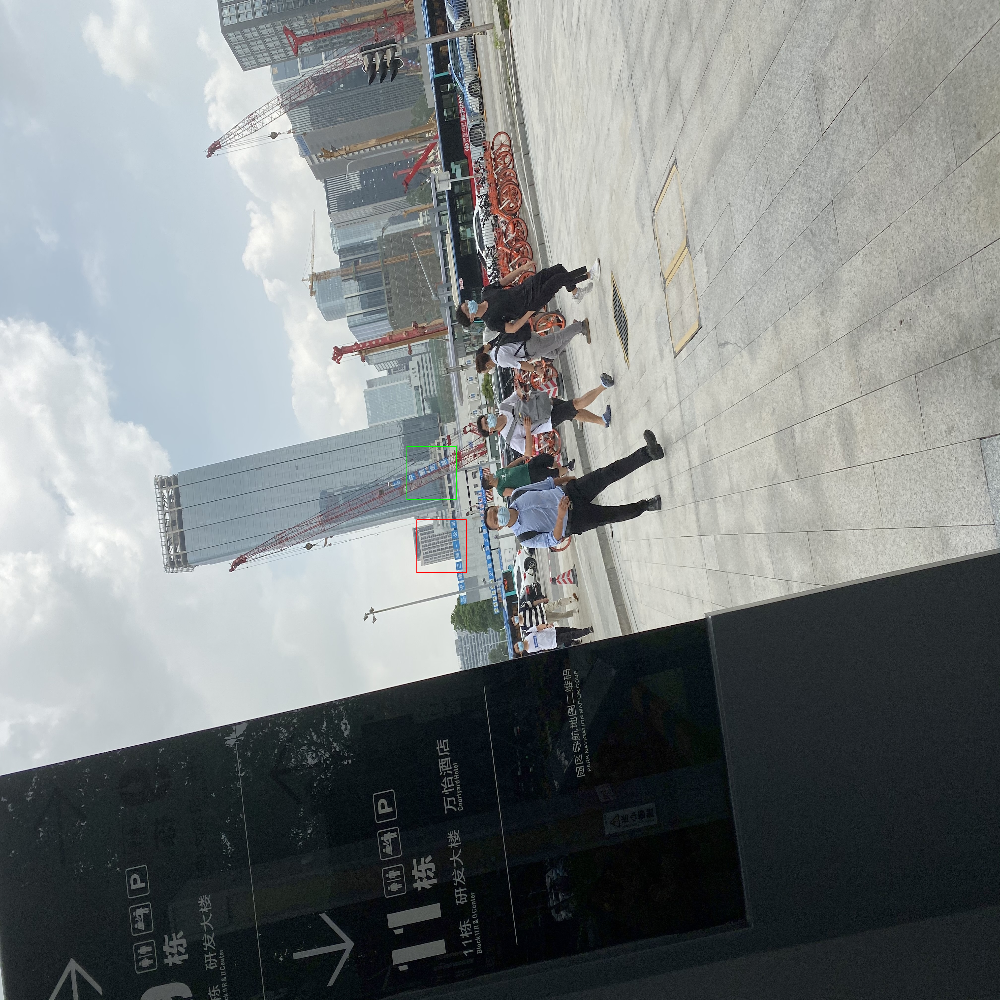}}\hspace{4pt}
\end{minipage}
\begin{minipage}[ht]{.99\linewidth}
\centering
\subfloat{\label{}\includegraphics[width=0.11\linewidth]{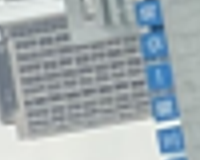}}\hspace{4pt}
\subfloat{\label{}\includegraphics[width=0.11\linewidth]{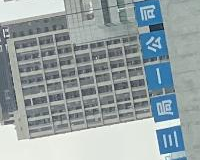}}\hspace{4pt}
\subfloat{\label{}\includegraphics[width=0.11\linewidth]{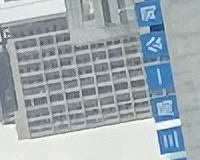}}\hspace{4pt}
\subfloat{\label{}\includegraphics[width=0.11\linewidth]{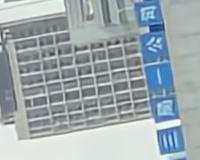}}\hspace{4pt}
\subfloat{\label{}\includegraphics[width=0.11\linewidth]{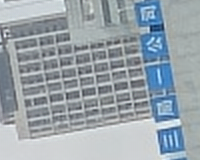}}\hspace{4pt}
\subfloat{\label{}\includegraphics[width=0.11\linewidth]{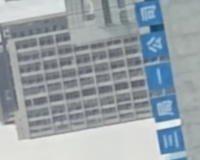}}\hspace{4pt}
\subfloat{\label{}\includegraphics[width=0.11\linewidth]{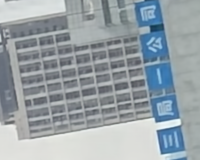}}\hspace{4pt}
\subfloat{\label{}\includegraphics[width=0.11\linewidth]{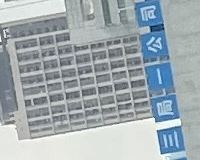}}\hspace{4pt}
\end{minipage}
\begin{minipage}[ht]{.99\linewidth}
\centering
\subfloat[Input wide-angle]{\label{}\includegraphics[width=0.11\linewidth]{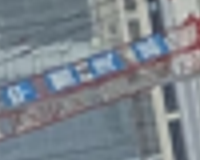}}\hspace{4pt}
\subfloat[Input telephoto]{\label{}\includegraphics[width=0.11\linewidth]{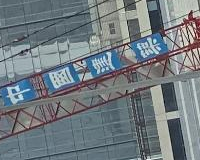}}\hspace{4pt}
\subfloat[DCSR]{\label{}\includegraphics[width=0.11\linewidth]{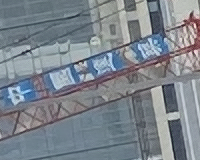}}\hspace{4pt}
\subfloat[DZSR]{\label{}\includegraphics[width=0.11\linewidth]{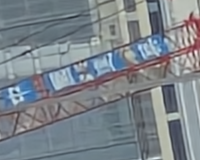}}\hspace{4pt}
\subfloat[ZEDUSR]{\label{}\includegraphics[width=0.11\linewidth]{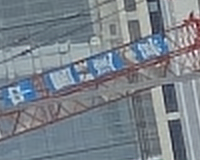}}\hspace{4pt}
\subfloat[Ours (efficient)]{\label{}\includegraphics[width=0.11\linewidth]{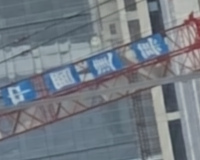}}\hspace{4pt}
\subfloat[Ours]{\label{}\includegraphics[width=0.11\linewidth]{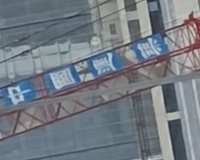}}\hspace{4pt}
\subfloat[GT]{\label{}\includegraphics[width=0.11\linewidth]{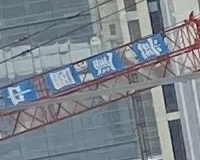}}\hspace{4pt}
\end{minipage}
\vspace{-5pt}
\caption{Example results in dual-camera super resolution on CameraFusion. Please see more results in supplementary materials.}
\label{fig:dual}
\end{figure*}

\captionsetup[subfloat]{labelsep=none,format=plain,labelformat=empty}
\begin{figure*}
\vspace{-5pt}
\begin{minipage}[ht]{.99\linewidth}
\centering
\subfloat{\label{}\includegraphics[width=0.11\linewidth]{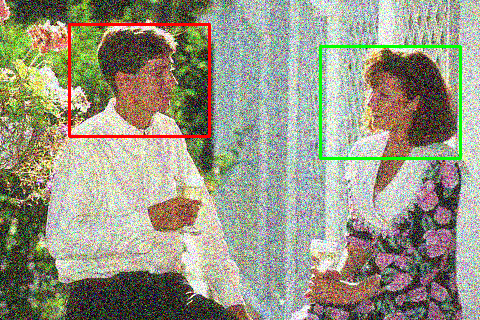}}\hspace{4pt}
\subfloat{\label{}\includegraphics[width=0.11\linewidth]{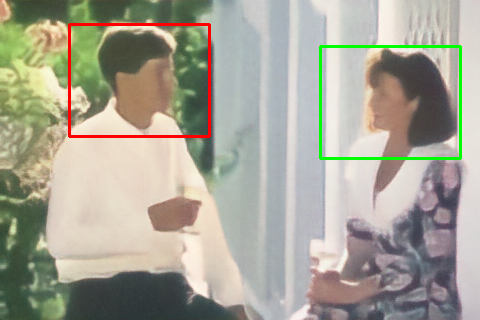}}\hspace{4pt}
\subfloat{\label{}\includegraphics[width=0.11\linewidth]{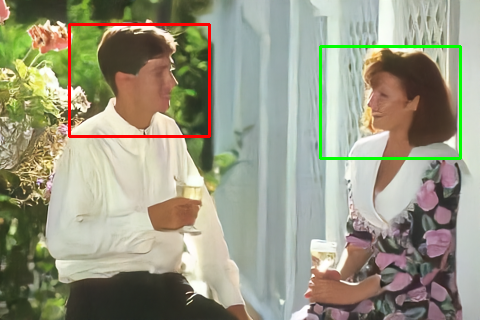}}\hspace{4pt}
\subfloat{\label{}\includegraphics[width=0.11\linewidth]{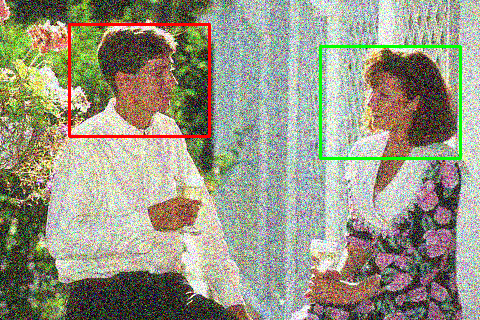}}\hspace{4pt}
\subfloat{\label{}\includegraphics[width=0.11\linewidth]{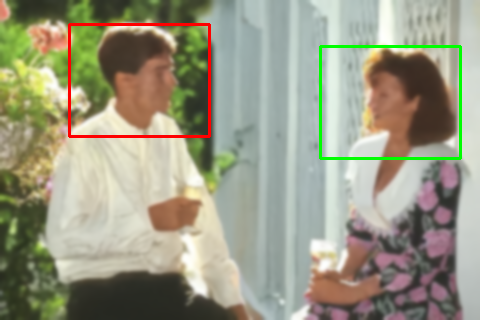}}\hspace{4pt}
\subfloat{\label{}\includegraphics[width=0.11\linewidth]{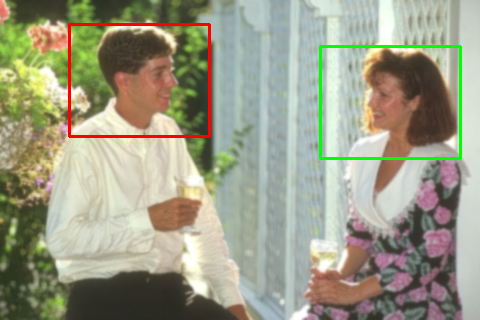}}\hspace{4pt}
\subfloat{\label{}\includegraphics[width=0.11\linewidth]{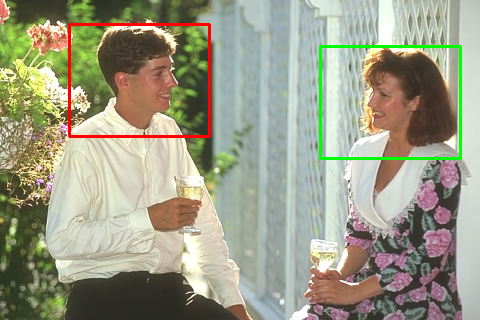}}\hspace{4pt}
\subfloat{\label{}\includegraphics[width=0.11\linewidth]{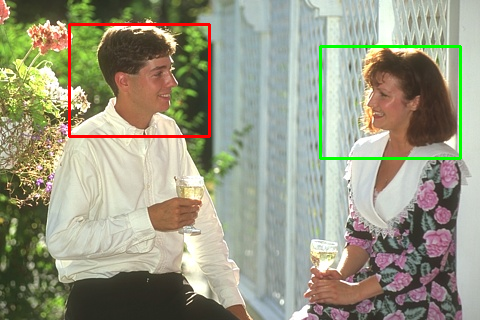}}\hspace{4pt}
\end{minipage}
\begin{minipage}[ht]{.99\linewidth}
\centering
\subfloat{\label{}\includegraphics[width=0.11\linewidth]{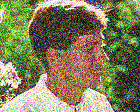}}\hspace{4pt}
\subfloat{\label{}\includegraphics[width=0.11\linewidth]{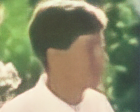}}\hspace{4pt}
\subfloat{\label{}\includegraphics[width=0.11\linewidth]{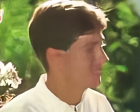}}\hspace{4pt}
\subfloat{\label{}\includegraphics[width=0.11\linewidth]{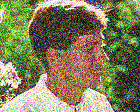}}\hspace{4pt}
\subfloat{\label{}\includegraphics[width=0.11\linewidth]{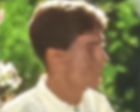}}\hspace{4pt}
\subfloat{\label{}\includegraphics[width=0.11\linewidth]{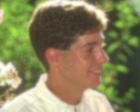}}\hspace{4pt}
\subfloat{\label{}\includegraphics[width=0.11\linewidth]{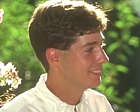}}\hspace{4pt}
\subfloat{\label{}\includegraphics[width=0.11\linewidth]{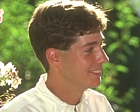}}\hspace{4pt}
\end{minipage}
\begin{minipage}[ht]{.99\linewidth}
\centering
\subfloat[Input]{\label{}\includegraphics[width=0.11\linewidth]{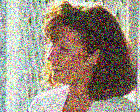}}\hspace{4pt}
\subfloat[MAXIM]{\label{}\includegraphics[width=0.11\linewidth]{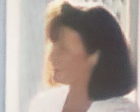}}\hspace{4pt}
\subfloat[PromptIR]{\label{}\includegraphics[width=0.11\linewidth]{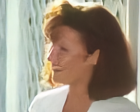}}\hspace{4pt}
\subfloat[KBNet]{\label{}\includegraphics[width=0.11\linewidth]{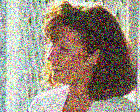}}\hspace{4pt}
\subfloat[DiffUIR]{\label{}\includegraphics[width=0.11\linewidth]{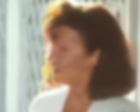}}\hspace{4pt}
\subfloat[Ours (efficient)]{\label{}\includegraphics[width=0.11\linewidth]{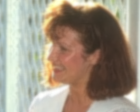}}\hspace{4pt}
\subfloat[Ours]{\label{}\includegraphics[width=0.11\linewidth]{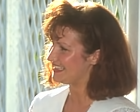}}\hspace{4pt}
\subfloat[GT]{\label{}\includegraphics[width=0.11\linewidth]{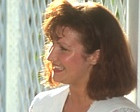}}\hspace{4pt}
\end{minipage}
\vspace{-5pt}
\caption{Example results in image denoising on Urban100. Please see more results in supplementary materials.}
\label{fig:denoise_}
\end{figure*}

\captionsetup[subfloat]{labelsep=none,format=plain,labelformat=empty}
\begin{figure*}
\vspace{-5pt}
\begin{minipage}[ht]{.99\linewidth}
\centering
\subfloat{\label{}\includegraphics[width=0.11\linewidth]{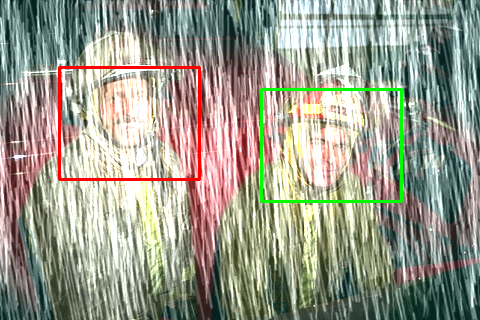}}\hspace{4pt}
\subfloat{\label{}\includegraphics[width=0.11\linewidth]{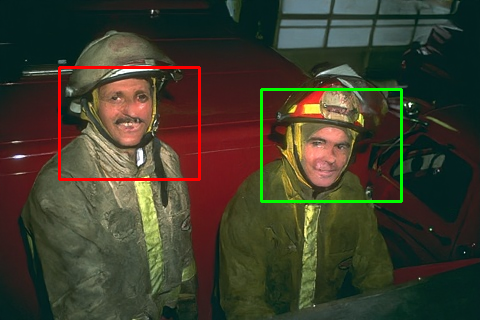}}\hspace{4pt}
\subfloat{\label{}\includegraphics[width=0.11\linewidth]{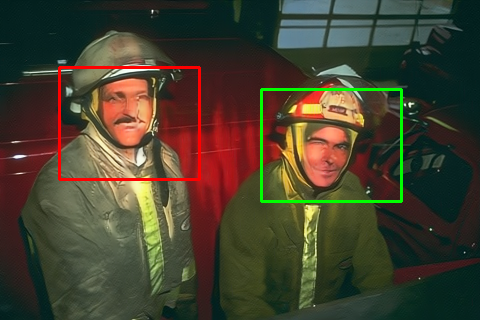}}\hspace{4pt}
\subfloat{\label{}\includegraphics[width=0.11\linewidth]{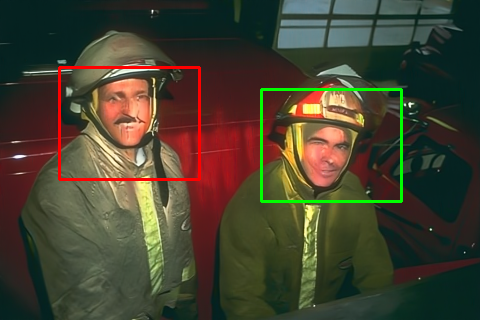}}\hspace{4pt}
\subfloat{\label{}\includegraphics[width=0.11\linewidth]{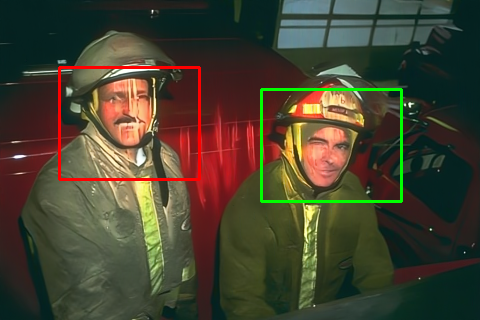}}\hspace{4pt}
\subfloat{\label{}\includegraphics[width=0.11\linewidth]{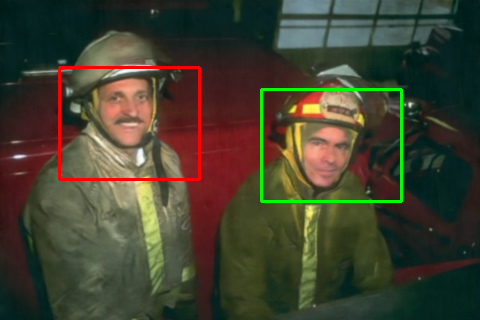}}\hspace{4pt}
\subfloat{\label{}\includegraphics[width=0.11\linewidth]{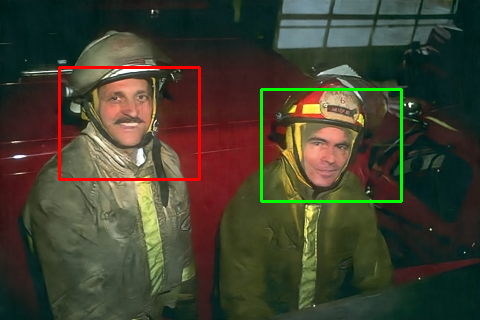}}\hspace{4pt}
\subfloat{\label{}\includegraphics[width=0.11\linewidth]{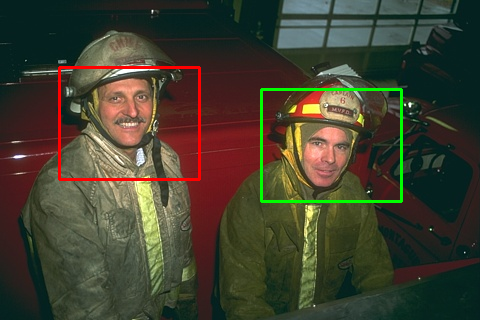}}\hspace{4pt}
\end{minipage}

\begin{minipage}[ht]{.99\linewidth}
\centering
\subfloat{\label{}\includegraphics[width=0.11\linewidth]{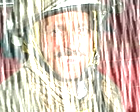}}\hspace{4pt}
\subfloat{\label{}\includegraphics[width=0.11\linewidth]{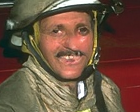}}\hspace{4pt}
\subfloat{\label{}\includegraphics[width=0.11\linewidth]{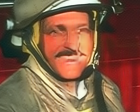}}\hspace{4pt}
\subfloat{\label{}\includegraphics[width=0.11\linewidth]{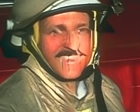}}\hspace{4pt}
\subfloat{\label{}\includegraphics[width=0.11\linewidth]{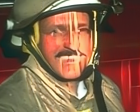}}\hspace{4pt}
\subfloat{\label{}\includegraphics[width=0.11\linewidth]{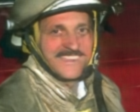}}\hspace{4pt}
\subfloat{\label{}\includegraphics[width=0.11\linewidth]{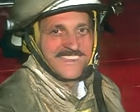}}\hspace{4pt}
\subfloat{\label{}\includegraphics[width=0.11\linewidth]{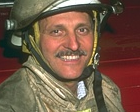}}\hspace{4pt}
\end{minipage}

\begin{minipage}[ht]{.99\linewidth}
\centering
\subfloat[Input]{\label{}\includegraphics[width=0.11\linewidth]{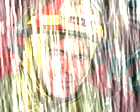}}\hspace{4pt}
\subfloat[DA-CLIP]{\label{}\includegraphics[width=0.11\linewidth]{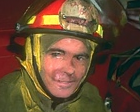}}\hspace{4pt}
\subfloat[MAXIM]{\label{}\includegraphics[width=0.11\linewidth]{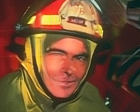}}\hspace{4pt}
\subfloat[Restormer]{\label{}\includegraphics[width=0.11\linewidth]{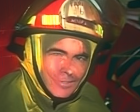}}\hspace{4pt}
\subfloat[KBNet]{\label{}\includegraphics[width=0.11\linewidth]{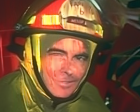}}\hspace{4pt}
\subfloat[Ours (efficient)]{\label{}\includegraphics[width=0.11\linewidth]{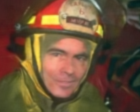}}\hspace{4pt}
\subfloat[Ours]{\label{}\includegraphics[width=0.11\linewidth]{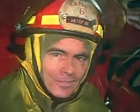}}\hspace{4pt}
\subfloat[GT]{\label{}\includegraphics[width=0.11\linewidth]{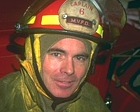}}\hspace{4pt}
\end{minipage}
\vspace{-5pt}
\caption{Example results in image deraining on Rain-100H. Please see more results in supplementary materials.}
\vspace{-10pt}
\label{fig:derain_}
\end{figure*}

\begin{table}[t]
\centering
\resizebox{0.5\textwidth}{!}{ 
\begin{tabular}{c|c|ccccc}
    \toprule
    Method & Dataset & PSNR (\(\uparrow\)) & SSIM (\(\uparrow\)) & LPIPS (\(\downarrow\)) & FID (\(\downarrow\)) & CLIPIQA (\(\uparrow\)) \\
    \midrule
    \textbf{No amplification} & RealSR & 22.74 & 0.7339 & 0.4115 & 104.67 & 0.1208 \\
    \textbf{Random amplification} & RealSR & 23.35 & 0.7368 & 0.3929 & 114.39 & \textbf{0.2136} \\
    \textbf{Ours (efficient)} & {RealSR} & \textbf{25.09} & \textbf{0.7430} & \textbf{0.3453} & \textbf{89.52} & 0.1899 \\
    \bottomrule
    \textbf{No amplification} & CameraFusion & 22.95 & 0.6614 & 0.3418 & 148.73 & 0.1498 \\
    \textbf{Random amplification} & CameraFusion & 27.61 & 0.7513 & 0.1979 & 61.08 & 0.1170 \\
    \textbf{Ours (efficient)} & CameraFusion & \textbf{28.68} & \textbf{0.7965} & \textbf{0.1767} & \textbf{59.85} & \textbf{0.1123} \\
    \bottomrule
\end{tabular}
} 
\caption{Ablation study about the efficient version of our method.}
\label{table:error}
\end{table}

\begin{table}[t]
\centering
\resizebox{0.5\textwidth}{!}{ 
\begin{tabular}{c|c|c|c}
    \toprule
    Memory/Time costs & ResShift (T=15) & Ours (T=15) & Ours efficient (T=15) \\
    \midrule
    Train (256*256) & 16.5GB/27h & 41.7GB/90h & 19.8GB/28h\\
    \bottomrule
    Test (256*256) & 3.6GB/0.96s & 3.6GB/0.96s & 3.6GB/0.96s\\ 
    \bottomrule
    Test (1024*1024) & 14.1GB/13s & 14.1GB/13s & 14.1GB/13s\\
    \bottomrule
\end{tabular}
} 
\caption{Computational costs of ResShift, our method, and our efficient version in training and testing stages. `h' and `s' are short for hours and seconds respectively.}
\vspace{-10pt}
\label{table:statis}
\end{table}

\section{Experiments}
\label{sec:experiments}
\subsection{Datasets and Implementation Details.}

\textbf{Datasets.}
We conduct experiments across five tasks: single image super-resolution (SISR), denoising, deraining, dehazing, and dual-camera super-resolution. For SISR experiment,
we adopt the 4x upscaling task. We utilize DIV2K \cite{agustsson2017ntire} and Flickr2K \cite{agustsson2017ntire} datasets for training, and RealSR \cite{ji2020real} dataset and CameraFusion \cite{wang2021dual} dataset for testing. For denoising, we follow \cite{potlapalli2306promptir,liang2021swinir} to leverage the training set of SIDD \cite{liu2009sidd} dataset for training, and Urban100 \cite{huang2015single} dataset for testing, where random Gaussian noise with a standard deviation of 50 is added to high-quality images to obtain the input images.  For deraining, we adopt Rain-13k \cite{jiang2020multi} dataset for training, and Rain-100H \cite{jiang2020multi} dataset for testing. For dehazing, we use the training set of RESIDE-6k \cite{li2018benchmarking} dataset for training, and the testing set of RESIDE-6k dataset for testing. For dual-camera super-resolution, we incorporate the training set of the CameraFusion dataset for training, and the testing set of CameraFusion dataset for testing. We follow ZEDUSR \cite{yue2024kedusr} to bicubic downsample the wide-angle images by a factor of 4 to obtain the input, while the original wide-angle images serve as the target images and the telephoto images act as references.

\textbf{Implementations of Our Method.}
Our method uses Resshift as the backbone DDM due to its efficiency. The ground-truth image ${\bf{GT}} = {\bf{x}}_{0}$, the input low-quality image is ${\bf{y}}_{0}$, and the residue ${\bf{r}}_0 = {\bf{y}}_{0}-{\bf{x}}_{0}$. At iteration $t$, the ground-truth image ${\bf{GT}}_t = {\bf{x}}_{0} + \eta_{t-1} {\bf{r}}_{0}$. In the forward process, the distortion addition operation at iteration $t$ is defined as $q({{\bf{x}}_t}|{{\bf{x}}_0})$, and $q({\bf{x}}_{t} | {\bf{x}}_{0}) = \mathcal{N}({\bf{x}}_{t}; {\bf{x}}_{0} + \eta_{t} {\bf{r}}_{0}, \kappa^{2} \eta_{t} {\bf{I}})$, where $\{\eta_{t}\}$ is a shifting sequence which monotonically increases with the iteration $t$, \(\kappa\) is a hyper-parameter, and \(\mathbf{I}\) is the identity matrix.

In the experiments, we use a single NVIDIA A6000 GPU with a total batch size of 4. We employ a cosine learning rate scheduler, setting a maximum learning rate of $1 \times 10^{-4}$. Additionally, we implement exponential moving average (EMA) to enhance training stability and overall model performance. All the training is carried out at a resolution of $256 \times 256$, with mean square error (MSE) as the metric $\beta_t$ of the loss function.

\textbf{Compasion Methods.}
We prioritize comparing the SOTA general restoration models, including the CNN based methods of AirNet, the transformer based methods of DA-CLIP, MAXIM, PromptIR, Restormer and SwinIR, and the DDM based methods of DiffPlugin and DiffUIR. In addition, we also compare the SOTA models for each task. For SISR, we compare Real-ESRGAN, HAT, Resshift, SeeSR, SRFormer and DiffIR. For denoising, we compare Blind2UnBlind, Pretrained-IPT and SADNet. For deraining, we compare Pretrained-IPT, WGWS and KBNet. For dehazing, we compare mixDehazeNet, dehazeFormer and WGWS. Dual-camera super-resolution is a special task, and we compare SwinIR, ZEDUSR, C$^{2}$-Matching, DCSR, DZSR, EDSR, MASA, RCAN, Real-ESRGAN, SRGAN, SRNTT and TTSR. We retrain the models based on the author's provided code if there are no released model weights. 

\captionsetup[subfloat]{labelsep=none,format=plain,labelformat=empty}
\begin{figure*}
\vspace{-5pt}
\begin{minipage}[ht]{.99\linewidth}
\centering
\subfloat{\label{}\includegraphics[width=0.11\linewidth]{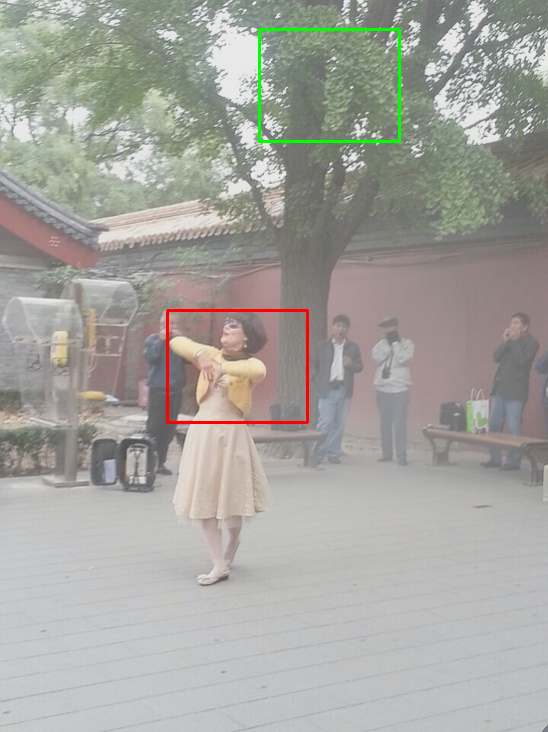}}\hspace{4pt}
\subfloat{\label{}\includegraphics[width=0.11\linewidth]{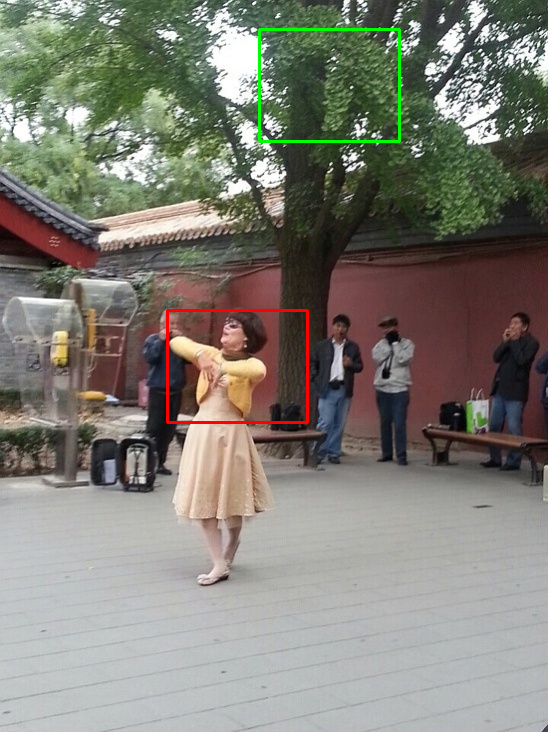}}\hspace{4pt}
\subfloat{\label{}\includegraphics[width=0.11\linewidth]{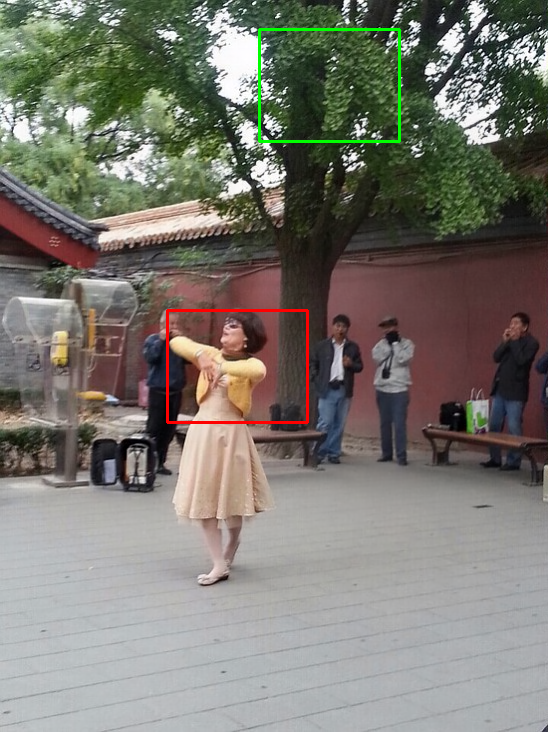}}\hspace{4pt}
\subfloat{\label{}\includegraphics[width=0.11\linewidth]{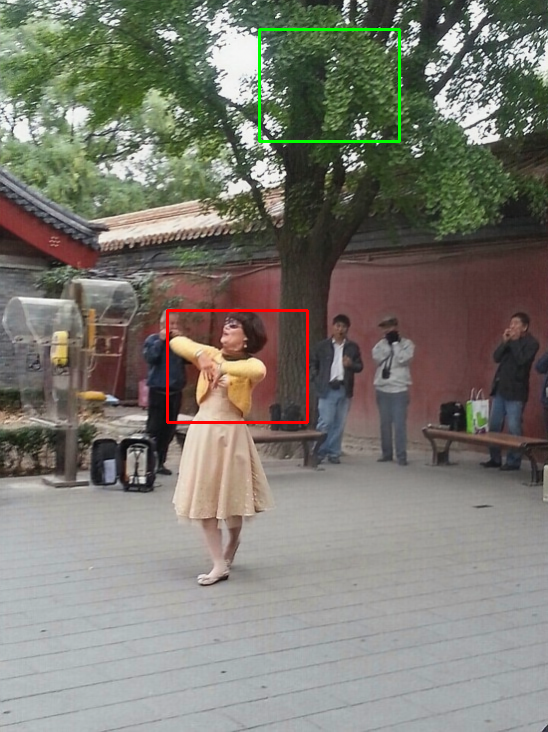}}\hspace{4pt}
\subfloat{\label{}\includegraphics[width=0.11\linewidth]{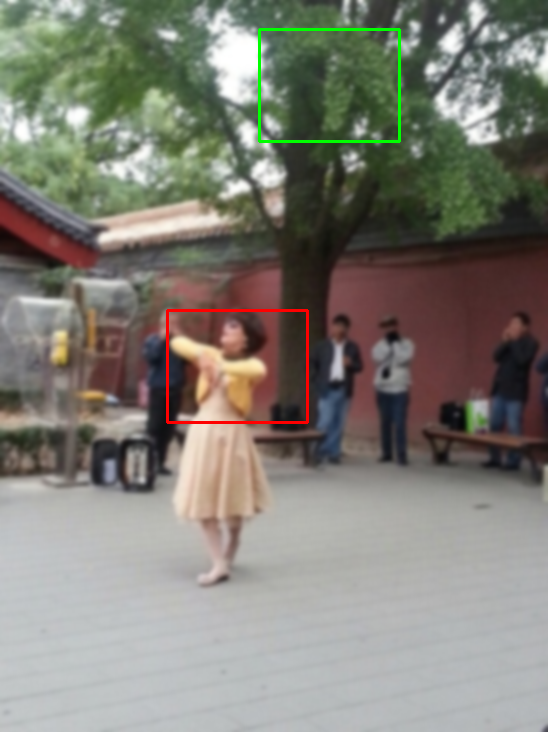}}\hspace{4pt}
\subfloat{\label{}\includegraphics[width=0.11\linewidth]{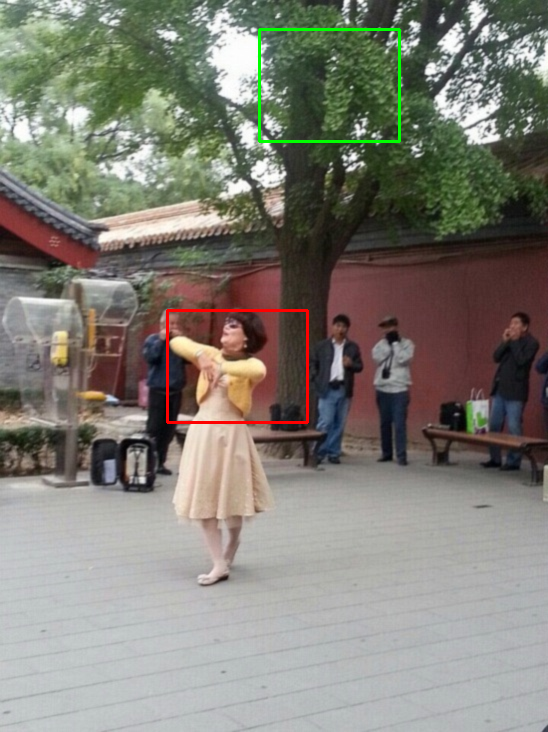}}\hspace{4pt}
\subfloat{\label{}\includegraphics[width=0.11\linewidth]{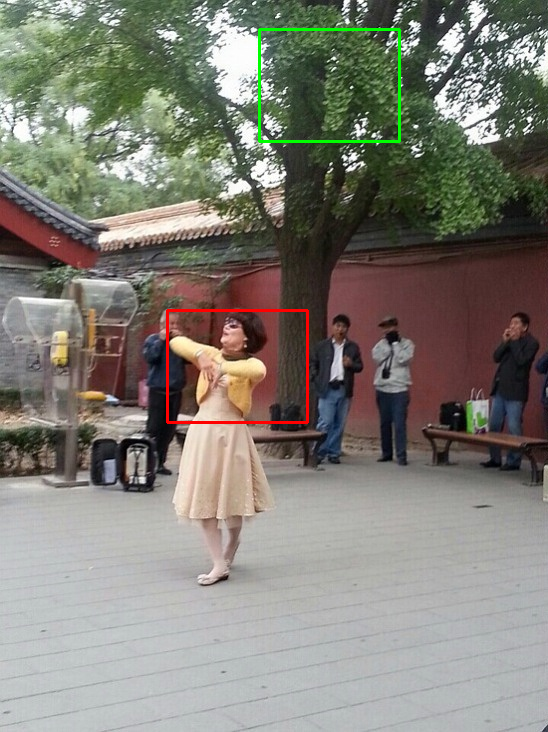}}\hspace{4pt}
\subfloat{\label{}\includegraphics[width=0.11\linewidth]{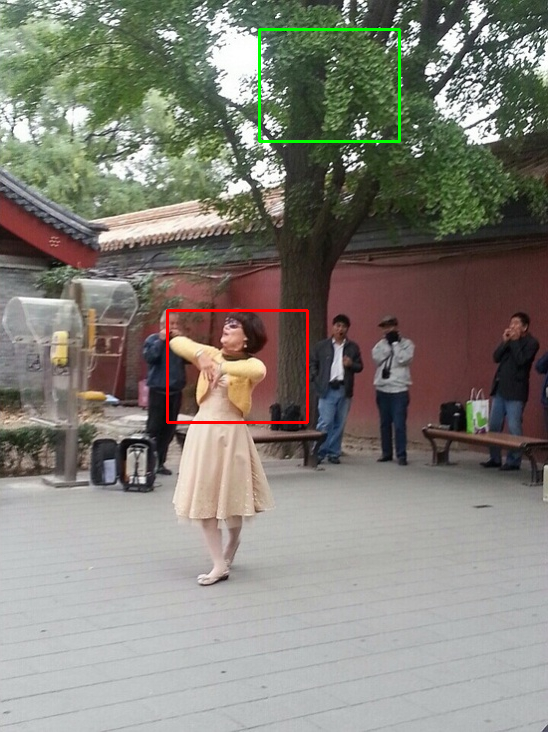}}\hspace{4pt}
\end{minipage}

\begin{minipage}[ht]{.99\linewidth}
\centering
\subfloat[Input]{\label{}\includegraphics[width=0.11\linewidth]{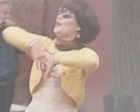}}\hspace{4pt}
\subfloat[DA-CLIP]{\label{}\includegraphics[width=0.11\linewidth]{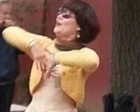}}\hspace{4pt}
\subfloat[MAXIM]{\label{}\includegraphics[width=0.11\linewidth]{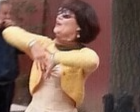}}\hspace{4pt}
\subfloat[Restormer]{\label{}\includegraphics[width=0.11\linewidth]{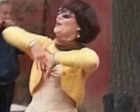}}\hspace{4pt}
\subfloat[PromptIR]{\label{}\includegraphics[width=0.11\linewidth]{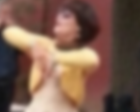}}\hspace{4pt}
\subfloat[Ours (efficient)]{\label{}\includegraphics[width=0.11\linewidth]{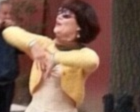}}\hspace{4pt}
\subfloat[Ours]{\label{}\includegraphics[width=0.11\linewidth]{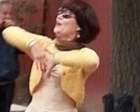}}\hspace{4pt}
\subfloat[GT]{\label{}\includegraphics[width=0.11\linewidth]{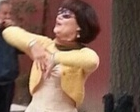}}\hspace{4pt}
\end{minipage}

\vspace{-5pt}
\caption{Example results in image dehaze on RESIDE-6k. Please see more results in supplementary materials.}

\label{fig:dehaze_}
\end{figure*}

\captionsetup[subfloat]{labelsep=none,format=plain,labelformat=empty}
\begin{figure*}
\vspace{-5pt}
\begin{minipage}[h]{.99\linewidth}
\centering
\subfloat{\label{}\includegraphics[width=0.13\linewidth]{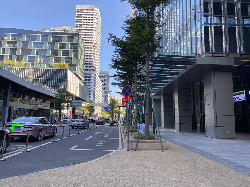}}\hspace{4pt}
\subfloat{\label{}\includegraphics[width=0.13\linewidth]{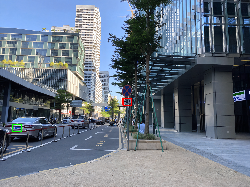}}\hspace{4pt}
\subfloat{\label{}\includegraphics[width=0.13\linewidth]{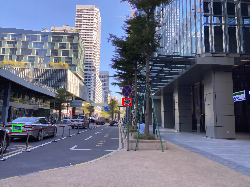}}\hspace{4pt}
\subfloat{\label{}\includegraphics[width=0.13\linewidth]{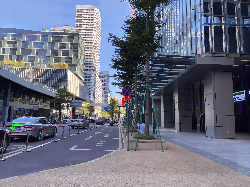}}\hspace{4pt}
\subfloat{\label{}\includegraphics[width=0.13\linewidth]{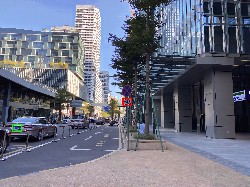}}\hspace{4pt}
\subfloat{\label{}\includegraphics[width=0.13\linewidth]{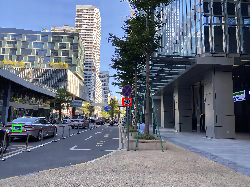}}\hspace{4pt}
\subfloat{\label{}\includegraphics[width=0.13\linewidth]{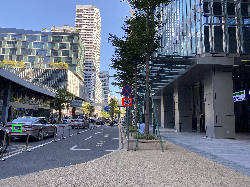}}\hspace{4pt}
\end{minipage}
\begin{minipage}[ht]{.99\linewidth}
\centering
\subfloat{\label{}\includegraphics[width=0.13\linewidth]{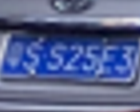}}\hspace{4pt}
\subfloat{\label{}\includegraphics[width=0.13\linewidth]{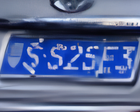}}\hspace{4pt}
\subfloat{\label{}\includegraphics[width=0.13\linewidth]{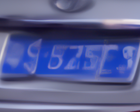}}\hspace{4pt}
\subfloat{\label{}\includegraphics[width=0.13\linewidth]{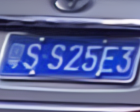}}\hspace{4pt}
\subfloat{\label{}\includegraphics[width=0.13\linewidth]{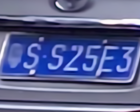}}\hspace{4pt}
\subfloat{\label{}\includegraphics[width=0.13\linewidth]{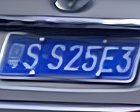}}\hspace{4pt}
\subfloat{\label{}\includegraphics[width=0.13\linewidth]{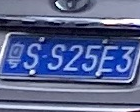}}\hspace{4pt}
\end{minipage}
\begin{minipage}[ht]{.99\linewidth}
\centering
\subfloat[Input]{\label{}\includegraphics[width=0.13\linewidth]{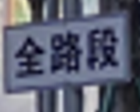}}\hspace{4pt}
\subfloat[ResShift]{\label{}\includegraphics[width=0.13\linewidth]{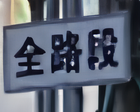}}\hspace{4pt}
\subfloat[LDM]{\label{}\includegraphics[width=0.13\linewidth]{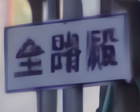}}\hspace{4pt}
\subfloat[$L_{nll}+L_{reg}$
]{\label{}\includegraphics[width=0.13\linewidth]{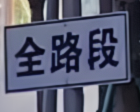}}\hspace{4pt}
\subfloat[LDM+DCT]{\label{}\includegraphics[width=0.13\linewidth]{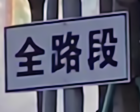}}\hspace{4pt}
\subfloat[ResShift+DCT]{\label{}\includegraphics[width=0.13\linewidth]{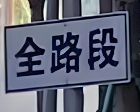}}\hspace{4pt}
\subfloat[GT]{\label{}\includegraphics[width=0.13\linewidth]{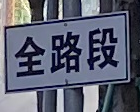}}\hspace{4pt}
\end{minipage}
\vspace{-5pt}
\caption{Example results of the ablation study about our data-consistent training. `DCT' is short for the proposed data-consistent training.}
\vspace{-10pt}
\label{fig:ablation_vis}
\end{figure*}

\subsection{Results and analysis.}
The quantitative results of five tasks are shown in Tables \ref{table:denoise}, \ref{table:derain}, \ref{table:dehaze}, \ref{table:dual} and \ref{table:sisr}. The qualitative results of five tasks are shown in Figs. \ref{fig:sisr1}, \ref{fig:sisr2}, \ref{fig:dual}, \ref{fig:denoise_}, \ref{fig:derain_} and \ref{fig:dehaze_}. Our results not only achieve high naturalness quality but also show high fidelity quality to the input without shape and color distortions. In the five tasks, our method usually achieves the best or top-three accuracy across various metrics. In contrast, other compared algorithms exhibit greater variability in accuracy across different test datasets and metrics. This indicates that our algorithm not only delivers strong results but also demonstrates robust generalization capabilities. In dual-camera super-resolution, SOTA comparison methods like DCSR often require complex architectures with multiple sub-networks to address challenges such as motion alignment, occlusion, and tonal inconsistencies between the input and reference images. In contrast, our method achieves SOTA accuracy without relying on specialized subnet designs or additional pre-/post-processing steps. This superior performance highlights the robust learning capacity of our model, which not only effectively learns the mapping from low-quality to high-quality but also is capable of addressing a variety of compound problems without the need for task-specific architectural modifications.

\textbf{Ablation Study.}
We try different variants of the proposed method. (1) To evaluate the benefits of our data-consistent training vs. the traditional training for different DDM backbones, besides ResShift, we also use latent diffusion model (LDM) as the backbone. Their results with the traditional training are denoted as ResShift and LDM, and their results with our data-consistent training are denoted as ResShift+DCT and LDM+DCT. (2) The work \cite{li2023error} also notices the gap between modular error and cumulative error and propose to use $L_{nll}+L_{reg}$ to train the DDM backbone, where $L_{t}^{nll}=\beta_t(f_{\theta }({\bf{x}}_t^{forw}),{\bf{GT}}_{t})$ follows the traditional DDM, $L_{t}^{reg}=\beta_t(f_{\theta }({\bf{x}}_t^{back}),{\bf{x}}_{t-1}^{forw})$ $L_{nll}$ pushes the output the same as the image generated by forward process. We also use ResShift as the backbone.

We also try different possible variants about our efficient version. (1) We follow Li et al. \cite{li2023error} to use the forward process to generate ${\bf{x}}_{t+1}^{train}={\bf{x}}_{t+1}^{forw}$, and use two times $f_{\theta}$ to process it to get the output at iteration $t$ without error amplification, denoted as `No amplification’. (2) We also try to use a random value between 0 and $(T-t)$ to amplify the error, denoted as `Random amplification’.

The results in Tables \ref{table:other_ablation}, \ref{table:error} and Fig. \ref{fig:ablation_vis} show the advantage of our choices and the generalizability of the proposed data-consistent training for different DDM backbones.

\section{Conclusions}

We address the limitations of traditional DDMs in image restoration by proposing a data-consistent training method to avoid shape and color distortions of the results. This method aligns the input data across both training and testing stages, ensuring that errors accumulated during testing are taken into consideration during training. Using ResShift as our backbone DDM, we achieve state-of-the-art accuracy on five popular image restoration tasks while maintaining high fidelity. Our method provides a general training solution for DDMs across various restoration tasks.

\clearpage

{\small
\bibliographystyle{ieeenat_fullname}
\bibliography{egbib}
}
\end{document}